\documentclass[Afour,sageh,times]{sagej}

\usepackage{moreverb,url}
\usepackage{lipsum}
\usepackage{subfig}
\usepackage{graphicx}
\usepackage{afterpage}
\usepackage{siunitx}
\usepackage{multirow}
\usepackage{tabularx}
\usepackage{pifont}
\usepackage{array}

\usepackage[colorlinks,bookmarksopen,bookmarksnumbered,citecolor=red,urlcolor=red]{hyperref}

\usepackage{color}
\definecolor{darkred}{rgb}{0.6,0.0,0.0}
\definecolor{darkgreen}{rgb}{0,0.50,0}
\definecolor{lightblue}{rgb}{0.0,0.42,0.91}
\definecolor{orange}{rgb}{0.99,0.48,0.13}
\definecolor{grass}{rgb}{0.18,0.80,0.18}
\definecolor{pink}{rgb}{0.97,0.15,0.45}

\usepackage{listings}

\lstdefinestyle{colored}{ %
  basicstyle=\ttfamily,
  backgroundcolor=\color{white},
  commentstyle=\color{green}\itshape,
  keywordstyle=\color{blue}\bfseries\itshape,
  stringstyle=\color{red},
}
\lstdefinelanguage{PythonPlus}[]{Python}{
  morekeywords=[1]{,as,assert,nonlocal,with,yield,self,True,False,None,} 
  morekeywords=[2]{,__init__,__add__,__mul__,__div__,__sub__,__call__,__getitem__,__setitem__,__eq__,__ne__,__nonzero__,__rmul__,__radd__,__repr__,__str__,__get__,__truediv__,__pow__,__name__,__future__,__all__,}, 
  morekeywords=[3]{,object,type,isinstance,copy,deepcopy,zip,enumerate,reversed,list,set,len,dict,tuple,range,xrange,append,execfile,real,imag,reduce,str,repr,}, 
  morekeywords=[4]{,Exception,NameError,IndexError,SyntaxError,TypeError,ValueError,OverflowError,ZeroDivisionError,}, 
  morekeywords=[5]{,ode,fsolve,sqrt,exp,sin,cos,arctan,arctan2,arccos,pi, array,norm,solve,dot,arange,isscalar,max,sum,flatten,shape,reshape,find,any,all,abs,plot,linspace,legend,quad,polyval,polyfit,hstack,concatenate,vstack,column_stack,empty,zeros,ones,rand,vander,grid,pcolor,eig,eigs,eigvals,svd,qr,tan,det,logspace,roll,min,mean,cumsum,cumprod,diff,vectorize,lstsq,cla,eye,xlabel,ylabel,squeeze,}, 
}
\lstdefinelanguage{PyBrIM}[]{PythonPlus}{
  emph={d,E,a,Fc28,Fy,Fu,D,des,supplier,Material,Rectangle,PyElmt},
}
\lstdefinestyle{colorEX}{
  basicstyle=\footnotesize\ttfamily,
  backgroundcolor=\color{white},
  commentstyle=\color{darkgreen}\slshape,
  keywordstyle=\color{blue}\bfseries\itshape,
  keywordstyle=[2]\color{blue}\bfseries,
  keywordstyle=[3]\color{grass},
  keywordstyle=[4]\color{red},
  keywordstyle=[5]\color{orange},
  stringstyle=\color{darkred},
  emphstyle=\color{pink}\underbar,
}

\newcommand\BibTeX{{\rmfamily B\kern-.05em \textsc{i\kern-.025em b}\kern-.08em
T\kern-.1667em\lower.7ex\hbox{E}\kern-.125emX}}

\def\journalname{The International Journal of Robotics Research}
\def\volumeyear{2025}

\newcommand{\cmark}{\ding{51}}%
\newcommand{\xmark}{\ding{55}}%
\newcommand{\ignore}[1]{}

\newcommand{\mc}[1]{\mathcal{#1}}

\newcommand{\bma}[1]{\left[\begin{array}{ #1}}
\newcommand{\ema}{\end{array}\right]}

\DeclareMathAlphabet{\mbf}{OT1}{ptm}{b}{n}
\newcommand{\mbs}[1]{{\boldsymbol{#1}}}

\def\fdotb{{\raisebox{-0.6ex}{ \kern0.2ex\raisebox{0.8ex}{\tiny $\hspace*{-1ex}\circ$}}}}
\def\fddotb{{\raisebox{-0.6ex}{ \kern0.2ex\raisebox{0.8ex}{\tiny $\hspace*{-1ex}\circ\circ$}}}}

\newcommand{\f}{\frac}

\newcommand{\trans}{{\ensuremath{\mathsf{T}}}} 
\newcommand{\utimes}{ {\raisebox{-0.6ex}{ \kern-1.0ex\raisebox{0.6ex}{ \small $\mathsf{v}$}}} } %
\newcommand{\beq}{\begin{equation}}
\newcommand{\eeq}{\end{equation}}
\newcommand{\bdis}{\begin{displaymath}}
\newcommand{\edis}{\end{displaymath}}
\newcommand{\beqarray}{\begin{eqnarray}}
\newcommand{\eeqarray}{\end{eqnarray}}
\newcommand{\beqarraynn}{\begin{eqnarray*}}
\newcommand{\eeqarraynn}{\end{eqnarray*}}
\newcommand{\balign}{\begin{align}}
\newcommand{\ealign}{\end{align}}
\newcommand{\balignnn}{\begin{align*}}
\newcommand{\ealignnn}{\end{align}}

\makeatletter
\renewcommand{\p@enumii}{\theenumi.}
\makeatother

\usepackage{arydshln}
\usepackage{mathtools}

\usepackage{amsmath} 
\usepackage{amssymb}
\usepackage{float}

\newcommand{\jln}[1]{\textcolor{black}{#1}}

\newcommand{\sh}[1]{\textcolor{black}{#1}}

\newcommand{\ijrr}[1]{\textcolor{black}{#1}}

\begin{document}

\runninghead{Mohammed Shalaby et al.}

\title{MILUV: A Multi-UAV Indoor Localization dataset with UWB and Vision}

\author{Mohammed Ayman Shalaby\affilnum{1}, Syed Shabbir Ahmed\affilnum{1}, Nicholas Dahdah\affilnum{1}, Charles Champagne Cossette\affilnum{1}, Jerome Le Ny\affilnum{2}, and James Richard Forbes\affilnum{1}}

\affiliation{\affilnum{1}Department of Mechanical Engineering, McGill University, Montreal, QC H3A 0C3, Canada.\\
\affilnum{2}Department of Electrical Engineering, Polytechnique Montreal, Montreal, QC H3T 1J4, Canada.}

\corrauth{Mohammed A. Shalaby, McGill University, Montreal, QC H3A 0C3, Canada.}

\email{mohammed.shalaby@mail.mcgill.ca}

\begin{abstract}
This paper introduces MILUV, a Multi-UAV Indoor Localization dataset with UWB and Vision measurements. This dataset comprises $217$ \sh{minutes} of flight time over \sh{$36$} experiments using three quadcopters, collecting ultra-wideband (UWB) ranging data such as the raw timestamps and channel-impulse response data, vision data from a stereo camera and a bottom-facing monocular camera, inertial measurement unit data, height measurements from a laser rangefinder, magnetometer data, and ground-truth poses from a motion-capture system. The UWB data is collected from up to $12$ transceivers affixed to mobile robots and static tripods in both line-of-sight and non-line-of-sight conditions. The UAVs fly at a maximum speed of $\SI{4.418}{\meter / \s}$ in an indoor environment with visual fiducial markers as features. \sh{MILUV is versatile and can be used for a wide range of applications beyond localization, but the primary purpose of MILUV is for testing and validating multi-robot UWB- and vision-based localization algorithms.} The dataset can be downloaded at \url{https://doi.org/10.25452/figshare.plus.28386041.v1}. A development kit is presented alongside the MILUV dataset, which includes benchmarking algorithms such as visual-inertial odometry, UWB-based localization using an extended Kalman filter, and classification of CIR data using machine learning approaches. The development kit can be found at \url{https://github.com/decargroup/miluv}\ijrr{, and is supplemented with a website available at \url{https://decargroup.github.io/miluv/}}.
\end{abstract}


\maketitle

\section{Introduction}
\sh{Autonomous vehicles rely} on the fusion of multiple complementary sensors to be able to perceive and navigate their environment. Sensor fusion is particularly critical in multi-robot scenarios, potentially aided by communication between robots to allow \jln{them} to perform tasks collaboratively, such as in infrastructure inspection, helping warehouse operators, or moving heavy objects over large distances. For example, autonomous road vehicles typically use a combination of radar, LiDAR, cameras, and global navigation satellite system (GNSS) receivers to localize themselves and map their environments. Unmanned aerial vehicles (UAVs), however, are more restricted in the sensors they can use, particularly when flying indoors. This has led to an increased interest in vision algorithms and the recent adoption of ultra-wideband (UWB) \jln{radios} to navigate in GNSS-denied environments. Cameras and UWB, unlike LiDAR sensors, are relatively inexpensive, low-power, and lightweight. Moreover, UWB provides a means of communication between multiple UAVs. 

Cameras and UWB sensors can be thought of as complementary. Cameras provide information about the environment and can be used for object detection and segmentation, including the detection of the relative pose of neighbouring robots. However, cameras are limited by their field-of-view (FOV) and require favourable lighting conditions. Meanwhile, UWB \jln{radios do not depend on} lighting conditions \jln{and, with appropriate antennas, are} largely omni-directional, so that they can provide information to robots outside a camera's FOV.
\jln{They can be used to exchange messages between transceivers over short distances with precise timestamps, 
as well as record the wireless channel state between the transceivers, in the form of channel impulse responses (CIRs).}
\jln{This information, although not as rich as for cameras, can then be used in particular to provide} 
range measurements to fixed \emph{anchors} with known positions and/or to neighbouring robots, \jln{from which
relative poses can also be estimated.}
However, \jln{although UWB is designed to be relatively robust to multipath propagation effects, 
these range measurements are still}
affected by non-line-of-sight (NLOS) conditions, when an object obstructs the direct path between 
the two communicating UWB transceivers. 

\begin{figure*}
    \centering
    \includegraphics[trim={0cm 1cm 0cm 1cm},clip,width=0.85\textwidth]{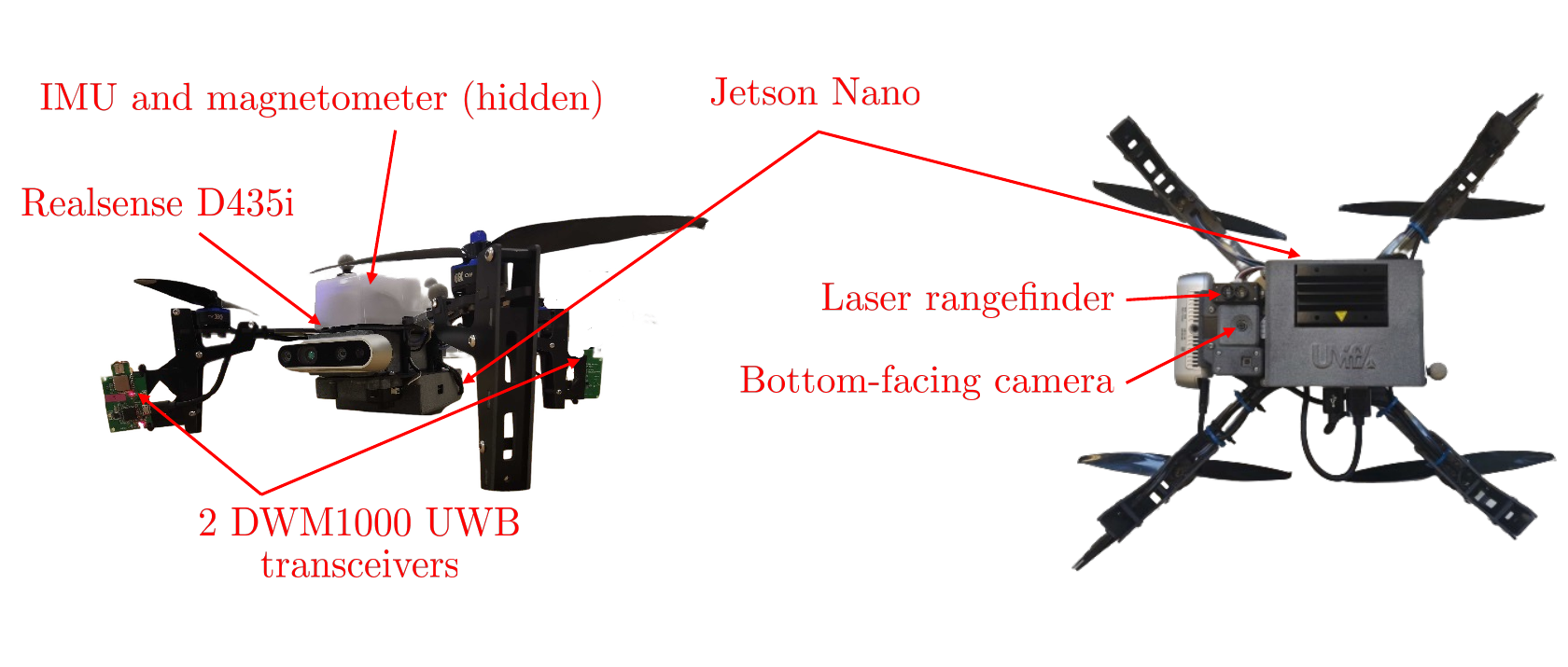}
    \caption{The sensors equipped on the Uvify IFO-S quadcopter.}
    \label{fig:ifo}
\end{figure*}

\begin{figure}[h!]
    \vspace{-0.5cm}
    \centering
    \subfloat{{\includegraphics[width = \columnwidth, trim={4cm 2cm 3cm 4cm}, clip]{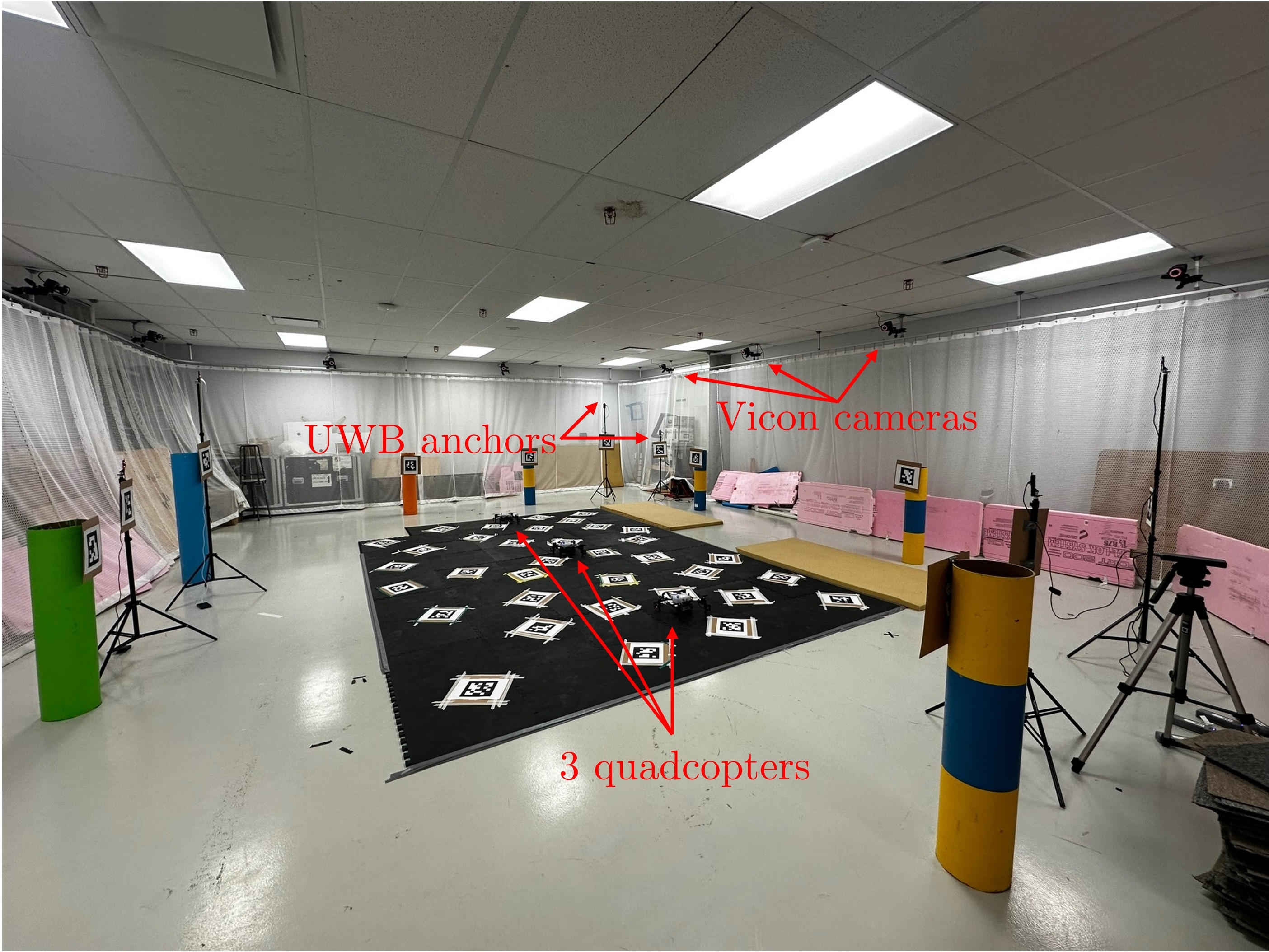}}} \\
    \subfloat{{\includegraphics[width = \columnwidth, trim={3cm 3cm 3cm 4cm}, clip]{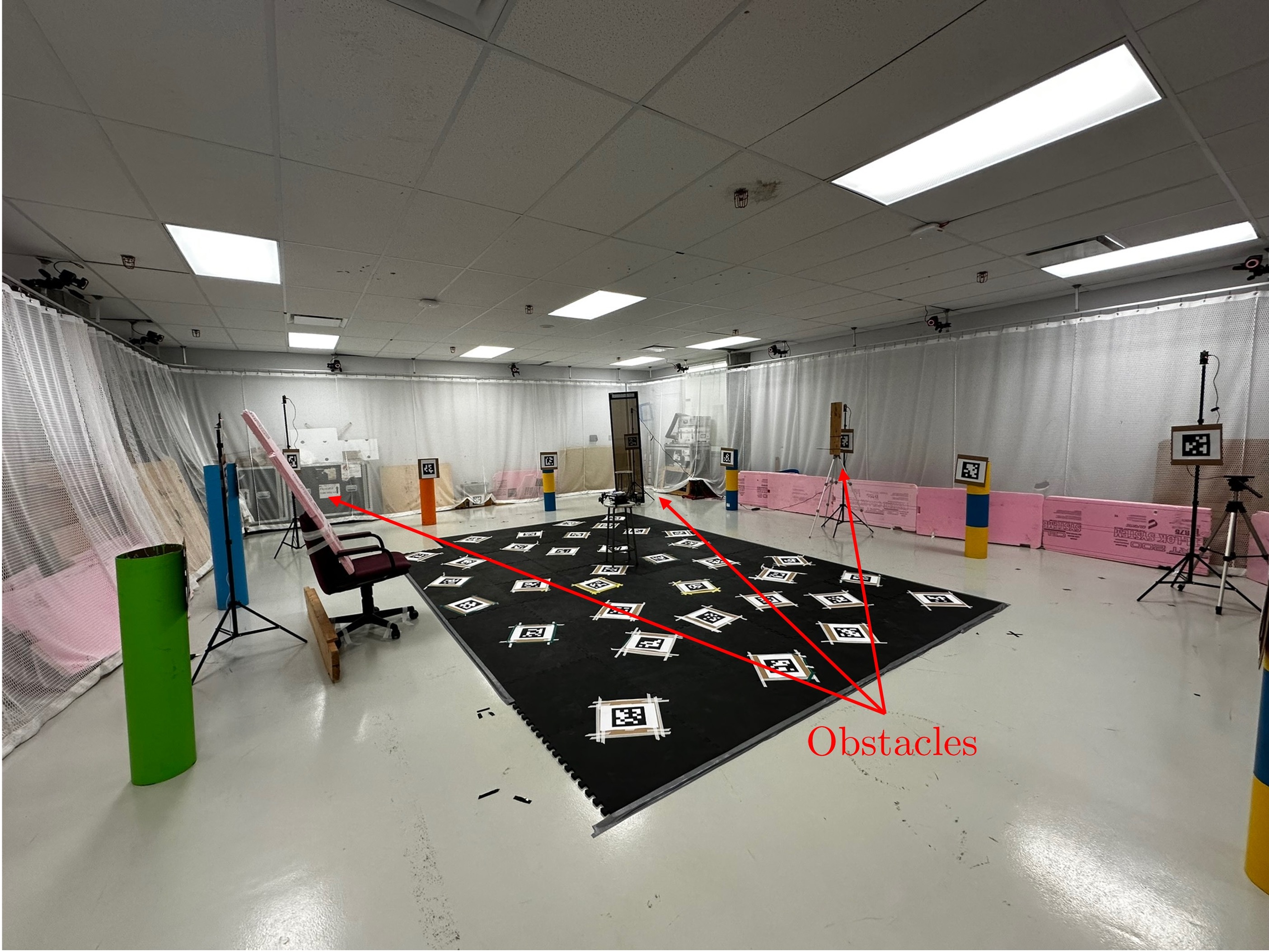}}}
    \caption{The environmental setup, with (top) and without (bottom) obstacles.}
    \label{fig:env}
\end{figure}

As such, the necessity to address indoor localization and mapping algorithms using vision and UWB simultaneously is evident. Utilizing vision and UWB for localization are two active fields of research where a lot of progress has been achieved, but the low-level complexities of each sensing modality are typically addressed separately and 
\jln{sensor information is}
only fused at a high level. 
To foster research at the intersection of these two actively growing fields, 
this paper presents MILUV, a Multi-UAV Indoor Localization dataset with UWB and Vision measurements. 
\jln{This dataset provides}
low-level UWB data such as raw timestamps, 
CIR data, and received signal power alongside measurements from a stereo camera, 
with the goal of supporting the development of advanced algorithms for multi-robot 
online localization and for camera and UWB calibration.

The \jln{MILUV} dataset includes data collected using three quadcopters, each equipped with \jln{several} sensors shown in Figure \ref{fig:ifo}, primarily a stereo camera and two UWB transceivers. The environment, shown in Figure \ref{fig:env}, is fitted with six UWB transceivers, for a total of twelve between anchors and robots. Alongside the cameras and the UWB transceivers, MILUV includes data from a downward-facing monocular camera, a laser rangefinder providing height measurements, a magnetometer, and two inertial measurement units (IMUs) on each quadcopter. A variety of different trajectories are \jln{recorded}, including static experiments and other special trajectories with and without obstacles intended for calibrating and characterizing sensor parameters. A different combination for the number of robots, the number of UWB transceivers per robot, and the UWB anchor constellation is used in each experiment. In total, $173\,\si{GB}$ of data is collected for a total of $217$ minutes of cumulative flight time. The dataset can be downloaded at \url{https://doi.org/10.25452/figshare.plus.28386041.v1}.

To aid with the usage of MILUV, the data has been benchmarked against standard estimation tools, including pose estimation by fusing the UWB and IMU measurements using an extended Kalman filter (EKF), visual-inertial simultaneous localization and mapping (SLAM) using VINS-Fusion \citep{Qin2019}, and fusing visual-inertial odometry (VIO) with UWB measurements in an EKF for pose estimation. Additionally, a development kit 
\jln{is provided}
at \url{https://github.com/decargroup/miluv} to allow users to parse the data in Python and to run the benchmark algorithms. \ijrr{Documentation on how to get started with the development kit is available at \url{https://decargroup.github.io/miluv/}}.

In summary, the main contributions of this dataset are as follows:
\begin{itemize}
    \item A UWB dataset with three quadcopters, each equipped with two UWB transceivers, a stereo camera, two IMUs, a downward-facing camera, a magnetometer, and a height laser rangefinder. Ground-truth pose data is collected for each quadcopter using a Vicon motion capture system.
    \item 
    Low-level UWB data \jln{recorded} with a double-sided two-way ranging (DS-TWR) \jln{protocol}, 
    time-synced with the vision data and all other sensors. This includes the raw timestamps, CIR data, received signal power, and clock skew measurements, in scenarios with and without obstacles for characterizing the effect of NLOS conditions.
    \item 
    Passive listening measurements \jln{recorded by UWB radios not participating in a given DS-TWR exchange}, 
    which provide information at no additional cost and can be used to develop new algorithms for multi-robot teams. 
\end{itemize}

\ijrr{Given the fact that MILUV exposes to the public a host of data that is typically underutilized in localization applications and robotics, an indoor setting is chosen for MILUV. There are many complications associated with outdoor settings with long sequences, including but not limited to obtaining accurate ground truth data, eliminating dynamic objects from the scenes, and ensuring consistent conditions between experiments. MILUV reduces the barrier to entry for the usage of low-level UWB data, and it does so by providing data in a simple setting with accurate ground truth data. This will contribute to the effort of bridging the gap between existing localization algorithms and ones that utilize more information available from UWB sensors, which is currently an understudied topic. The hope is that MILUV will foster further research in this direction, eventually bringing the field to a point where outdoor experimentation and testing using multiple vehicles will be feasible.} 

The rest of this paper is organized as follows. \jln{Related work is discussed} in Section \ref{sec:related_work}, and the notation used throughout is provided in Section \ref{sec:notation}. Details on the environmental setup and sensor setup are provided in Sections \ref{sec:environment} and \ref{sec:sensor_setup} respectively, and the sensor calibration procedure is discussed in Section \ref{sec:sensor_calibration}. The experiments provided in this dataset are summarized in Section \ref{sec:data_collection}, whereas the development kit and benchmarking results are provided in Section \ref{sec:dataset_usage}. Lastly, potential applications are discussed in Section \ref{sec:potential_applications}, and known limitations of this dataset are mentioned in Section \ref{sec:limitations_and_possible_extensions}.

\section{Related Work} \label{sec:related_work}
\begin{table*}[h]
\footnotesize\sf\centering
\renewcommand{\arraystretch}{1.4}
\caption{Public UWB and vision localization datasets relevant to MILUV. The datasets providing both UWB and vision data appear twice and are highlighted in blue.}
\begin{tabular}{ccccccccccc}
\toprule
\multicolumn{8}{c}{\textbf{UWB Localization Datasets}} \\
\midrule
Reference & Mobile Platform & Timestamps & CIR & Tags per robot & Anchors & Experiments & Obstacles\\
\midrule
\cite{Ledergerber2019} & 1 Roomba & \xmark & \cmark & 1 & 5 & 10 & \cmark \\
\cite{Queralta2020} & 1 quadcopter & \xmark & \xmark & 2 & 3-6 & 19 & \xmark  \\
\cite{Moron2023} & 1 quadcopter/5 turtlebots & \xmark & \xmark & 2 & 1-8 & 24 & \xmark \\
UTIL & \multirow{2}{*}{1 quadcopter} & \multirow{2}{*}{\xmark} & \multirow{2}{*}{\xmark} & \multirow{2}{*}{1} & \multirow{2}{*}{8} & \multirow{2}{*}{43} & \multirow{2}{*}{\cmark}  \\
\noalign{\vspace{-5pt}}\cite{Zhao2024} \\
\textcolor{blue}{NTU VIRAL} & \multirow{2}{*}{1 quadcopter} & \multirow{2}{*}{\xmark} & \multirow{2}{*}{\xmark} & \multirow{2}{*}{4} & \multirow{2}{*}{3} & \multirow{2}{*}{9} & \multirow{2}{*}{\xmark} \\
\noalign{\vspace{-5pt}}\cite{Nguyen2022} \\
\textcolor{blue}{MCD} & \multirow{2}{*}{1 ground vehicle/handheld rig} & \multirow{2}{*}{\xmark} & \multirow{2}{*}{\xmark} & \multirow{2}{*}{2} & \multirow{2}{*}{Unknown} & \multirow{2}{*}{17} & \multirow{2}{*}{\xmark}  \\
\noalign{\vspace{-5pt}}\cite{mcdviral2024} \\
\textcolor{blue}{STAR-Loc} & \multirow{2}{*}{1 handheld rig} & \multirow{2}{*}{\cmark} & \multirow{2}{*}{\xmark} & \multirow{2}{*}{2} & \multirow{2}{*}{8} & \multirow{2}{*}{21} & \multirow{2}{*}{\xmark}  \\
\noalign{\vspace{-5pt}}\cite{Dumbgen2023} \\
\textcolor{blue}{\textbf{MILUV}} & 3 quadcopters & \cmark & \cmark & 2 & 6 & 36 & \cmark \\
\midrule \midrule
\multicolumn{8}{c}{\textbf{Vision Localization Datasets}} \\
\midrule
Reference & Mobile Platform & \multicolumn{2}{c}{Features} & \multicolumn{2}{c}{Ground Truth} & Dimension & Indoor \\
\midrule
EuRoC & \multirow{2}{*}{1 quadcopter} & \multicolumn{2}{c}{\multirow{2}{*}{Industrial/indoors}} & \multicolumn{2}{c}{\multirow{2}{*}{Leica/mocap}} & \multirow{2}{*}{3D} & \multirow{2}{*}{\cmark}  \\
\noalign{\vspace{-5pt}}\cite{Burri2016} \\
Blackbird & \multirow{2}{*}{1 quadcopter} & \multicolumn{2}{c}{\multirow{2}{*}{Simulated/indoors}} & \multicolumn{2}{c}{\multirow{2}{*}{Mocap}} & \multirow{2}{*}{3D} & \multirow{2}{*}{\cmark} \\
\noalign{\vspace{-5pt}}\cite{Antonini2020} \\
Kimera-Multi & \multirow{2}{*}{8 ground vehicles} & \multicolumn{2}{c}{\multirow{2}{*}{Urban/corridors}} & \multicolumn{2}{c}{\multirow{2}{*}{GPS/Lidar}} & \multirow{2}{*}{2D} & \multirow{2}{*}{\cmark}  \\
\noalign{\vspace{-5pt}}\cite{Tian2023} \\
UZH-FPV & \multirow{2}{*}{1 quadcopter} & \multicolumn{2}{c}{\multirow{2}{*}{Hangar/field}} & \multicolumn{2}{c}{\multirow{2}{*}{Leica}} & \multirow{2}{*}{3D} & \multirow{2}{*}{\cmark}  \\
\noalign{\vspace{-5pt}}\cite{Delmerico2019} \\
\textcolor{blue}{NTU VIRAL} & \multirow{2}{*}{1 quadcopter} & \multicolumn{2}{c}{\multirow{2}{*}{Urban scenes}} & \multicolumn{2}{c}{\multirow{2}{*}{3D Laser Tracker}} & \multirow{2}{*}{3D} & \multirow{2}{*}{\xmark} \\
\noalign{\vspace{-5pt}}\cite{Nguyen2022} \\
\textcolor{blue}{MCD} & \multirow{2}{*}{1 ground vehicle/handheld rig} & \multicolumn{2}{c}{\multirow{2}{*}{Campus Areas}} & \multicolumn{2}{c}{\multirow{2}{*}{LiDAR/IMU/Survey Maps}} & \multirow{2}{*}{3D} & \multirow{2}{*}{\xmark} \\
\noalign{\vspace{-5pt}}\cite{mcdviral2024} \\
\textcolor{blue}{STAR-Loc} & \multirow{2}{*}{1 handheld rig} & \multicolumn{2}{c}{\multirow{2}{*}{April Tags}} & \multicolumn{2}{c}{\multirow{2}{*}{Mocap}} & \multirow{2}{*}{3D} & \multirow{2}{*}{\cmark} \\
\noalign{\vspace{-5pt}}\cite{Dumbgen2023} \\
\textcolor{blue}{\textbf{MILUV}} & 3 quadcopters & \multicolumn{2}{c}{April Tags} & \multicolumn{2}{c}{Mocap} & 3D & \cmark \\
\end{tabular}
\label{tab:datasets}
\end{table*}

A summary of relevant datasets in the literature is provided in Table \ref{tab:datasets}, showing an abundance of UWB and vision localization datasets that use a single mobile platform. These datasets typically focus on one sensing modality, and are therefore categorized as either UWB or Vision, with NTU VIRAL \citep{Nguyen2022}, MCD \citep{mcdviral2024}, STAR-Loc \citep{Dumbgen2023}, and MILUV being the exception. 

The publicly available datasets generally 
\jln{have only}
one mobile platform for data collection, such as \cite{Queralta2020} and UTIL \citep{Zhao2024} for UWB, and EuRoC \citep{Burri2016}, Blackbird \citep{Antonini2020}, and UZH-FPV \citep{Delmerico2019} for vision. \ijrr{MCD \citep{mcdviral2024} uses a quadcopter in some experiments, and a handheld rig in other experiments, and provides both UWB and vision data in large-scale outdoor campus areas.} However, the topic of multi-robot localization is growing in importance, \sh{where teams of robots} need to perform tasks collaboratively. As such, \cite{Tian2023} provides a comprehensive dataset with eight ground vehicles equipped with cameras and other sensors in different environments. Meanwhile, \cite{Moron2023} provide range measurements between five ground vehicles, each equipped with a single UWB transceiver. Nonetheless, to the authors' knowledge, there 
\jln{is} no multi-UAV dataset with UWB transceivers, despite UWB being a great fit 
for multi-UAV applications as it is a lightweight and low-power 
\jln{technology} that additionally provides a means for the robots to communicate. 

The majority of UWB localization datasets provide \jln{only} high-level 
UWB information, typically range measurements using some ranging protocol, 
such as two-way ranging (TWR) as in \cite{Queralta2020}, or time-difference-of-arrival (TDoA) 
as in UTIL \citep{Zhao2024}. For the localization datasets mentioned 
in Table \ref{tab:datasets}, only \cite{Ledergerber2019} and STAR-LOC \citep{Dumbgen2023} provide some lower-level UWB information, with the former providing CIR data and the latter providing the raw timestamps used to compute the range measurements. MILUV provides range measurements, the raw timestamps, CIR \jln{data}, 
as well as other low-level \sh{UWB} information such as the received signal power and 
passively-received messages, in order to 
develop and evaluate advanced localization algorithms that utilize 
more information made available by the UWB transceivers. 

Despite not providing low-level UWB information, NTU VIRAL \citep{Nguyen2022} is the dataset most similar to \sh{MILUV}, as it involves a UAV, includes multiple UWB tags per robot, and includes both UWB and vision information. NTU VIRAL provides an alternative to MILUV for a single-robot application outdoors, whereas MILUV \sh{goes beyond} NTU VIRAL by providing multiple ranging robots, providing low-level UWB information, and adding experiments with obstacles. The presence of April Tags and a motion capture system makes MILUV a \jln{good} candidate for the evaluation of localization algorithms. Also similar to MILUV is STAR-Loc \citep{Dumbgen2023}, where the same UWB hardware \sh{as in MILUV} is used. MILUV can be seen as the multi-robot extension of STAR-LOC, with additional sensors, additional UWB data, non-line-of-sight measurements, and more experiments.

\section{Notation} \label{sec:notation}
\begin{figure}
    \centering
    \includegraphics[trim={0cm 0cm 0cm 0cm},clip,width=0.95\columnwidth]{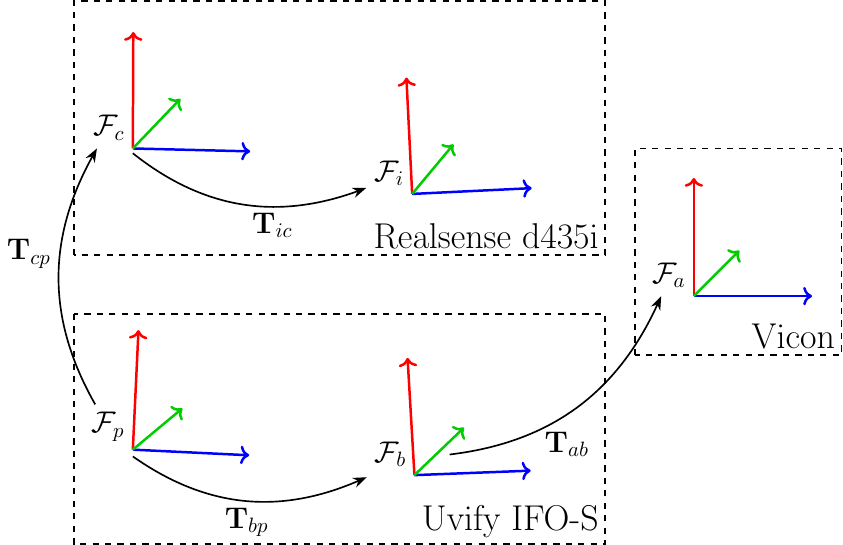}
    \caption{Sensor frames and their relative transforms.}
    \label{fig:frames}
\end{figure}

Throughout this paper, a bold upper-case letter such as $\mbf{X} \in \mathbb{R}^{m \times n}$ denotes real-valued $m \times n$ a matrix, and a bold lower-case letter such as $\mbf{x} \in \mathbb{R}^n$ denotes a real-valued $n \times 1$ column matrix. Meanwhile, a reference frame $a$ is denoted using $\mc{F}_a$. The five reference frames used in this dataset, 
shown in Figure \ref{fig:frames}, are 
\begin{itemize}
    \item an \emph{absolute} or world frame $\mc{F}_a$,
    \item a body-fixed frame $\mc{F}_b$,
    \item a camera-fixed frame $\mc{F}_c$, 
    \item an IMU frame for the Pixhawk IMU $\mc{F}_p$, and
    \item an IMU frame for the camera IMU $\mc{F}_i$.
\end{itemize}
Moreover, a vector resolved in reference frame $\alpha$ is denoted as $\mbf{x}_\alpha \in \mathbb{R}^n$, and the position vector of point $w$ relative to point $z$ as resolved in reference frame $\alpha$ is denoted as $\mbf{r}_\alpha^{wz} \in \mathbb{R}^3$. The direction cosine matrix (DCM), or rotation matrix, $\mbf{C}_{\alpha \beta} \in SO(3)$ relates the same physical vector resolved in two different frames such that $\mbf{x}_\alpha = \mbf{C}_{\alpha \beta} \mbf{x}_\beta$. The transformation matrix $\mbf{T}_{\alpha \beta} \in SE(3)$ is composed of the DCM $\mbf{C}_{\alpha \beta} \in SO(3)$ and the position $\mbf{r}^{wz}_\alpha \in \mathbb{R}^3$.

\section{Environment} \label{sec:environment}
\subsection{Flight Arena}
For the duration of the data collection, the quadcopters are within an approximate $\SI{4}{\meter} \times \SI{4}{\meter} \times \SI{3}{\meter}$ subsection of an enclosed flight arena. The arena has a Vicon motion capture system comprising 12 cameras. The experimental setup is depicted in Figure \ref{fig:env}.


\subsection{UWB Anchors}

\begin{figure*}
    \vspace{-0.5cm}
    \centering
    \subfloat[\centering Anchor Constellation~0]{{\includegraphics[width = 0.33\textwidth, trim={13cm 1cm 10cm 3cm}, clip]{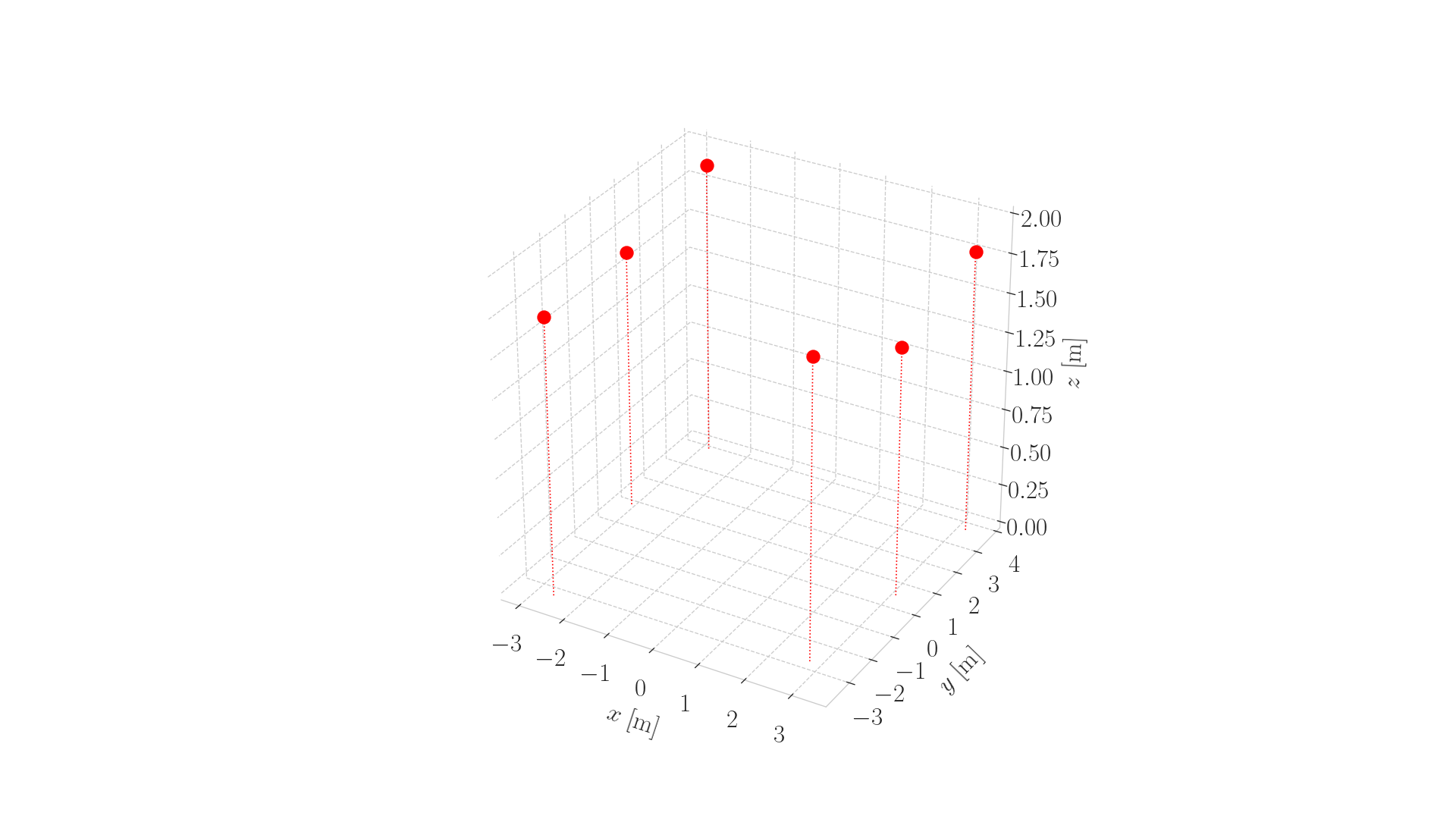}}}
    \subfloat[\centering Anchor Constellation~1]{{\includegraphics[width = 0.33\textwidth, trim={13cm 1cm 10cm 3cm}, clip]{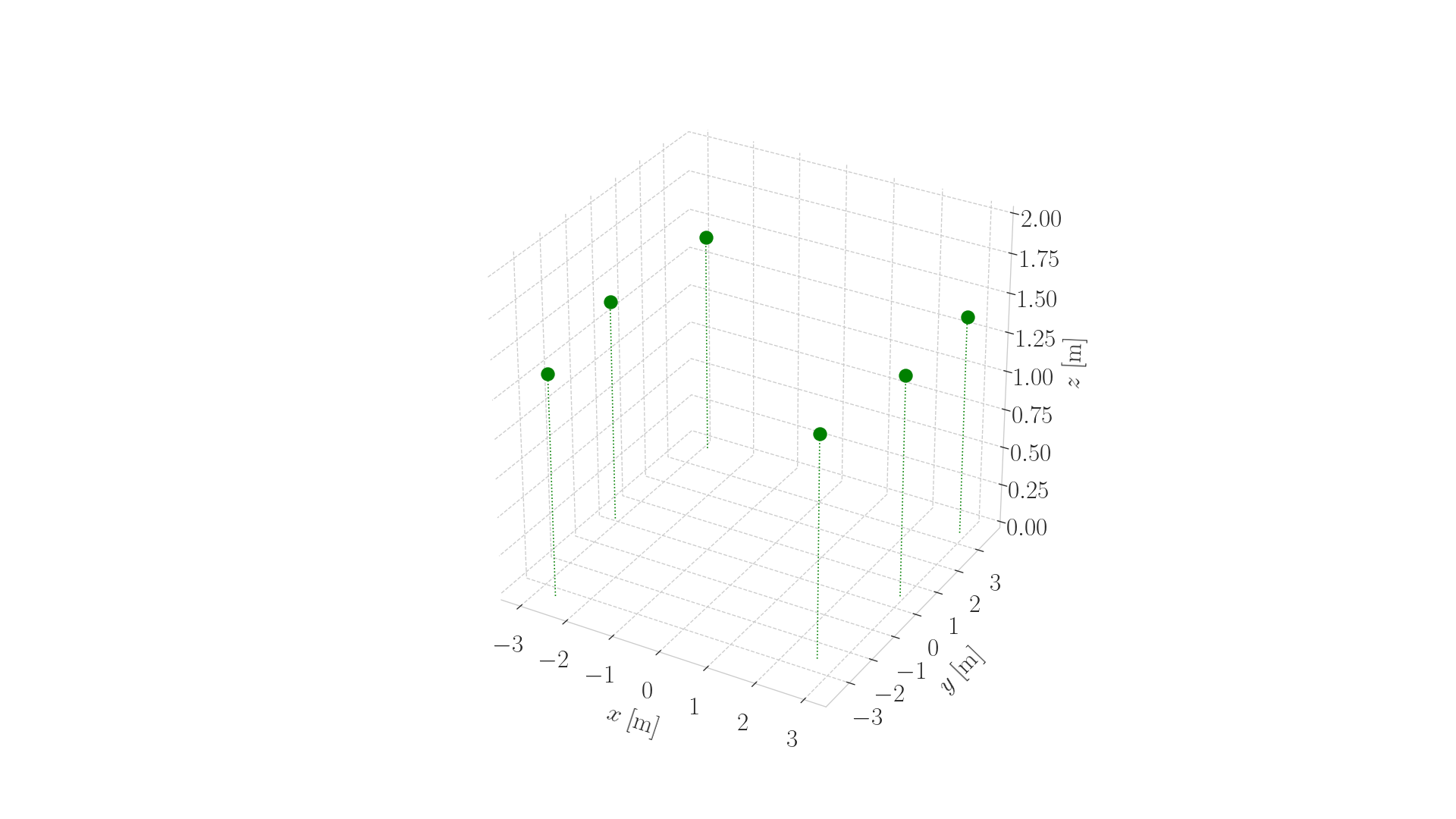}}}
    \subfloat[\centering Anchor Constellation~2]{{\includegraphics[width = 0.33\textwidth, trim={13cm 1cm 10cm 3cm}, clip]{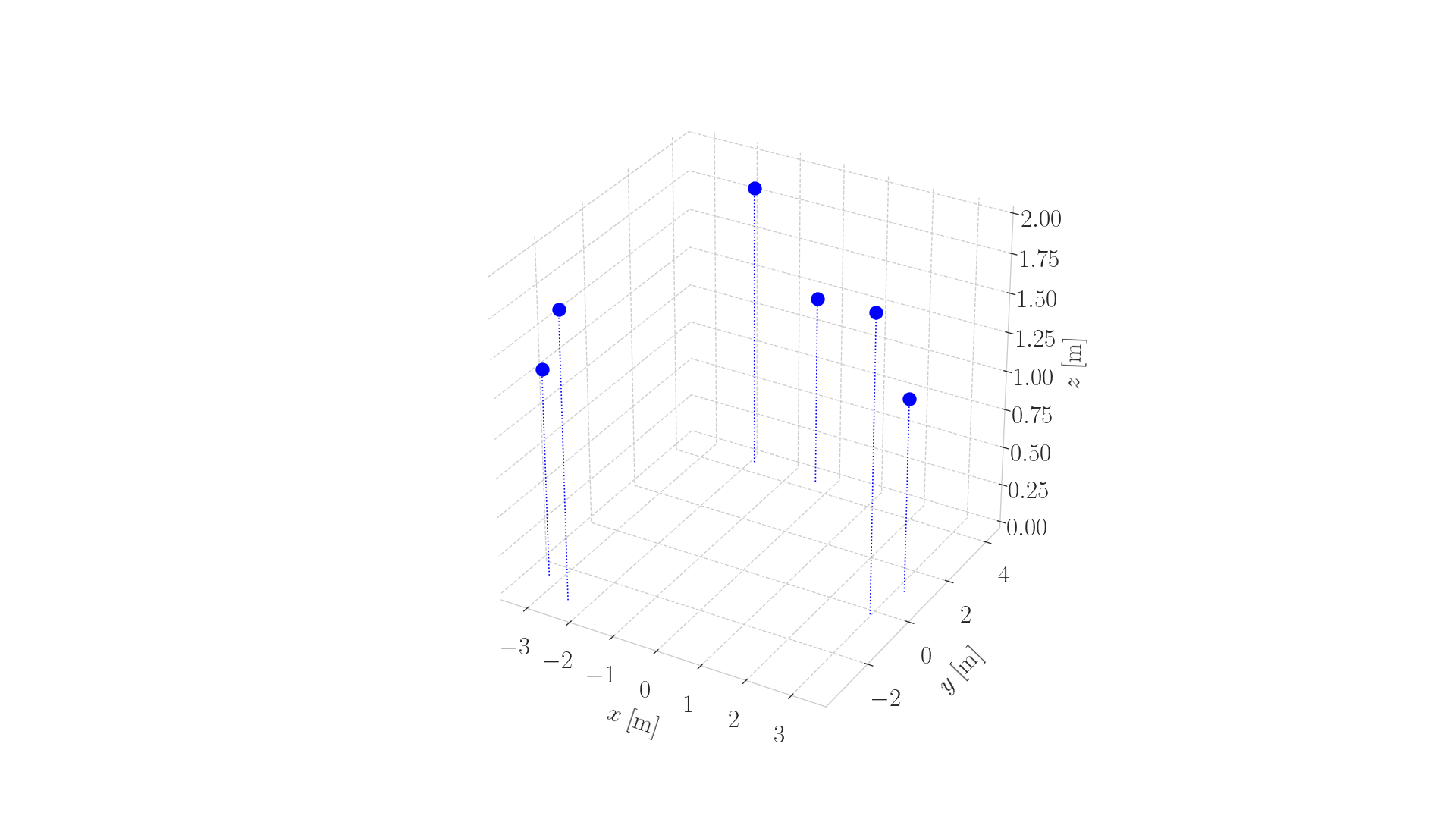}}}
    \caption{The three anchor constellations used in the \jln{data} collection of MILUV.}
    \label{fig:anchors}
\end{figure*}

The dataset includes experiments with 6 UWB anchors with anchor IDs 0 through 5, and with 3 different anchor constellations, shown in Figure~\ref{fig:anchors}. The primary constellation consists of anchors at varying heights, evenly spaced around the quadcopters' operating area. The second constellation consists of anchors at the same positions as the primary constellation, but with each transceiver at the same height. The third constellation consists of three clusters of two anchors at varied heights. The location of each anchor is determined using the Vicon motion capture system, \ijrr{and is provided in the development kit at \texttt{config/uwb/anchors.yaml} for the different constellations}. Wooden, acrylic plastic, and polystyrene obstacles are placed in front of UWB anchors \ijrr{4, 3, and 1, respectively}, as shown in Figure \ref{fig:env}, in order to disrupt line-of-sight (LOS) to the quadcopters for the duration of certain experiments.

\begin{figure}
\centering
    {\includegraphics[width = 0.75\columnwidth, trim={13cm 1cm 10cm 3cm}, clip]{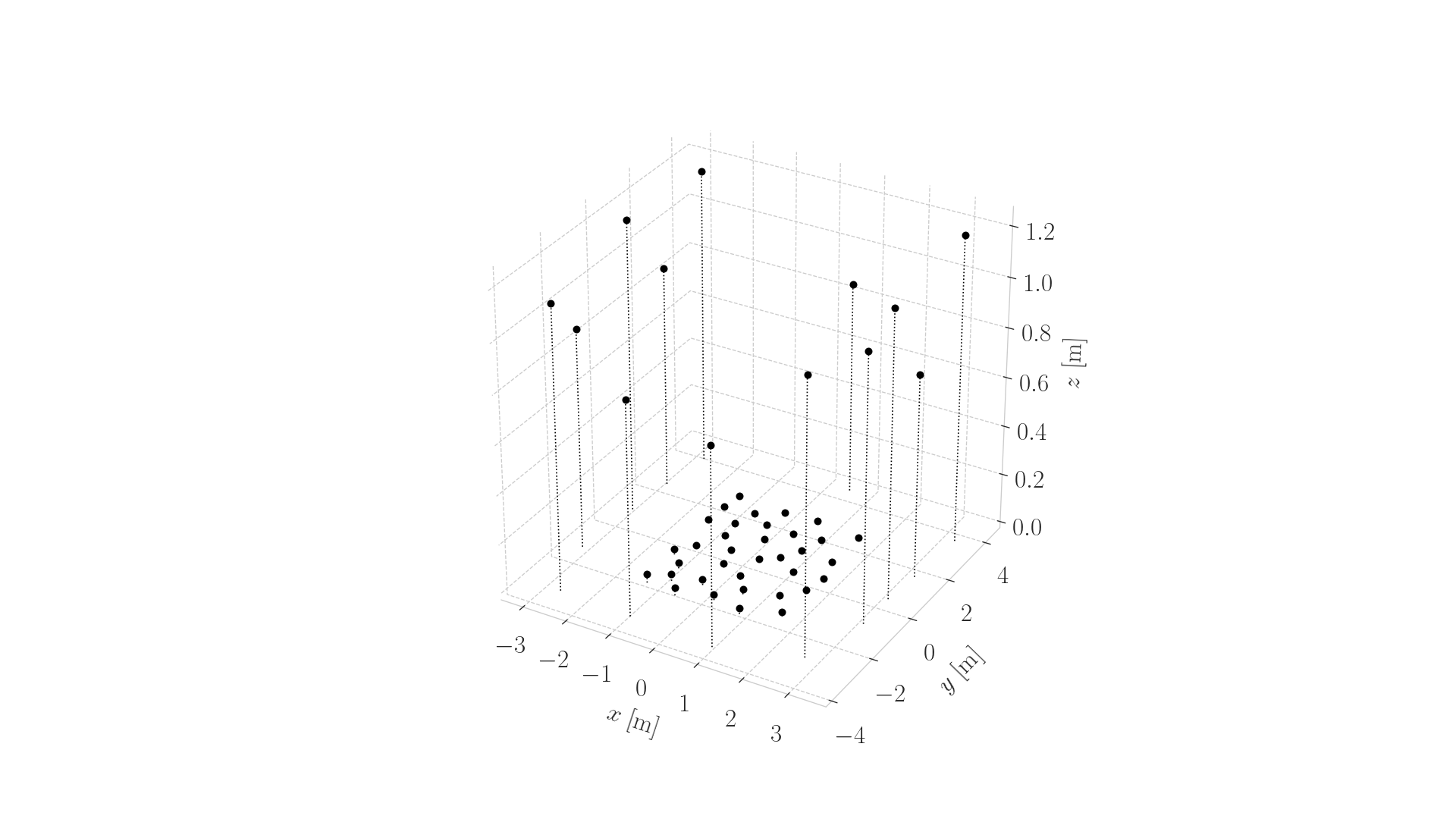}}
    \caption{\centering Positions of AprilTags used in the collection of this dataset.}
    \label{fig:april_tags}
\end{figure}

\subsection{Visual Features}

The arena is fitted with 48 36h11 AprilTags~\citep{Olson2011} with 35 tags affixed to the ground, 7 to static objects, and 1 below each of the 6 UWB anchors. The location of each AprilTag is determined using the Vicon motion capture system, and the results are shown in Figure~\ref{fig:april_tags}\ijrr{, and provided in the development kit at \texttt{config/apriltags/apriltags.yaml}}.


\section{Sensor Setup} \label{sec:sensor_setup}
\subsection{The Quadcopters}

The MILUV dataset comprises measurements from three Uvify IFO-S quadcopters. Each quadcopter is equipped with an IMU, a front-facing Intel Realsense D435i stereo camera, an integrated downward facing camera, and two UWB transceivers, depicted in Figure~\ref{fig:ifo}. The onboard computer is an NVIDIA Jetson Nano, and the flight controller is a Pixhawk 4 microcontroller running PX4 autopiloting software. The sensor frames and their relative transformations are illustrated in Figure~\ref{fig:frames}.

\subsection{Ultra-Wideband \jln{Transceivers}} 

A main feature of this dataset is the availability of low-level data available from the UWB transceivers, such as the raw timestamps, received signal power, clock skew measurements, and the channel impulse response. In this section, details regarding the custom hardware and firmware utilized for collecting the UWB data are discussed.

\subsubsection{Hardware}

\begin{figure}
    \centering
    \includegraphics[trim={0cm 0cm 0cm 0cm},clip,width=0.8\columnwidth]{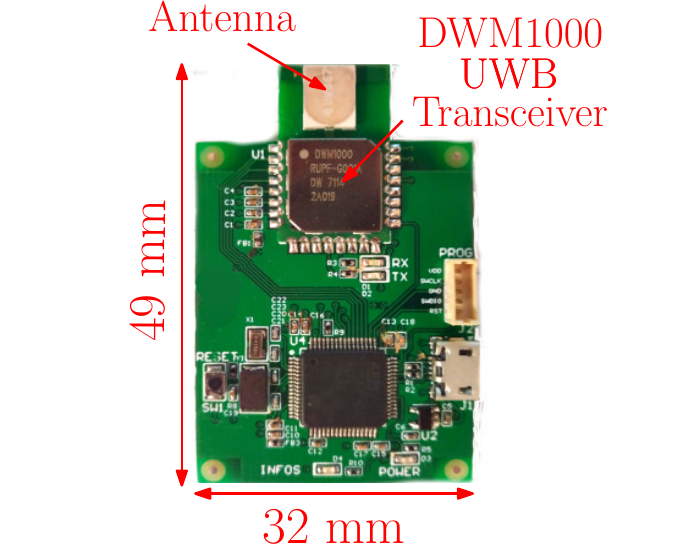}
    \caption{Custom-made board fitted with a DWM1000 UWB transceiver.}
    \label{fig:ourboard}
\end{figure}

Each UAV has two custom-made UWB modules mounted on diagonally opposing legs. These transceivers are labelled 10, 11, 20, 21, 30, and 31, with the first digit representing the robot to which each transceiver is attached, and the second digit differentiating the transceivers on an individual robot. The custom-built UWB boards, shown in Figure~\ref{fig:ourboard}, are built around the DWM1000\footnote{\url{https://www.qorvo.com/products/p/DW1000}} UWB transceiver, which is a commonly available and used transceiver. The use of the DWM1000 module with a custom-built board allows complete customizability and accessibility of the DWM1000 microprocessor's registers. Futhermore, as these boards will be placed on medium-sized quadcopters, they are designed to be compact and lightweight, measuring at 32~\si{mm}~$\times$~49~\si{mm} and weighing just $8~\si{grams}$.

\jln{Custom} firmware for these transceivers is written in \texttt{C} 
in a manner that allows easy integration with \emph{robot operating system} (ROS), which is used for the interface between the UWB transceivers and the other sensors and computers running on the quadcopters. The custom firmware further allows the implementation of custom ranging protocols and media-access control algorithms tailored to multi-robot localization and navigation.


Two UWB transceivers are fitted to every quadcopter approximately $45~\si{cm}$ apart, as shown in Figure~\ref{fig:ifo}, using a 3D-printed support. These mobile transceivers are hereinafter referred to as \emph{tags}. Placing two UWB tags per robot allows for better observability properties of pose estimators, as shown in \cite{Shalaby2021rpe}. 
The positions of all tags on the robots relative to the robots' body frames, 
\jln{measured} using the Vicon motion capture system,
are shown in Table~\ref{tab:moment_arms}. \ijrr{These are also provided in the development kit at \texttt{config/uwb/tags.yaml}}.

\begin{table}[h]
\small\sf\centering
\caption{The position vector of each tag on the robots 
\jln{(i.e., $\mbf{r}_i^{ji}$ for Tag $j$ on Robot $i$)}
relative to the origin of the body frame of the robot the tag is \jln{attached} to, 
each resolved in that same body frame. 
}
\begin{tabular}{ccc}
Robot ID& Tag ID &$\mbf{r}_i^{ji}$\\
\midrule
\multirow{2}{*}{ifo001}& 10 &$\bma{ccc}0.132 & -0.172 & -0.052\ema^\trans$\\
 &11 & $\bma{ccc}-0.175 & 0.157 & -0.053\ema^\trans$ \\
\midrule
\multirow{2}{*}{ifo002}& 20 & $\bma{ccc}0.165 & -0.151 & -0.035\ema^\trans$ \\
 &21 & $\bma{ccc}-0.155 & 0.170 & -0.017\ema^\trans$ \\
\midrule
\multirow{2}{*}{ifo003}& 30 & $\bma{ccc}0.167 & -0.181 & -0.056\ema^\trans$ \\
 &31 & $\bma{ccc}-0.135 & 0.155 & -0.052\ema^\trans$ \\
\end{tabular}
\label{tab:moment_arms}
\end{table}

\subsubsection{Ranging Protocol} \label{subsubsec:ranging_protocol}

\begin{figure}
    \centering
    \includegraphics[width=\columnwidth]{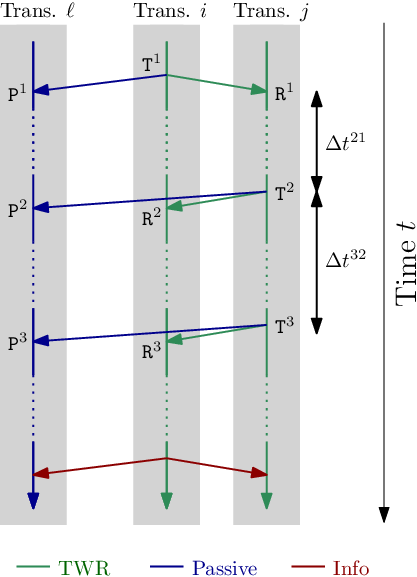}
    \caption{The DS-TWR protocol used in MILUV, showing an instance where Transceivers $i$ and $j$ are ranging 
    with each other and a Transceiver $\ell$ is passively listening on the messages. After every ranging transaction, an information message shown here in red is transmitted by the transaction initiator, here Transceiver $i$, to share timestamps $\mathtt{T}^1$, $\mathtt{R}^2$, and $\mathtt{R}^3$ with all other transceivers.}
    \label{fig:ranging_protocol}
\end{figure}

The ranging protocol refers to the sequence of transmissions and receptions between UWB transceivers to compute a range measurement. The ranging protocol used in MILUV is a variation of the standard double-sided 
two-way ranging (DS-TWR) protocol, based on \cite{shalaby2023calib}. 
As shown by the green arrows in Figure \ref{fig:ranging_protocol}, in a ranging transaction
between a pair of transceivers, an initiating transceiver transmits a message to a neighbouring 
transceiver, \jln{who responds with two messages.}
Both transceivers timestamp in their own clocks all message transmission and reception instances, where transmission and reception timestamps are denoted $\mathtt{T}^k$ and $\mathtt{R}^k$, respectively, for the $k^\text{th}$ message. A range measurement is then computed as 
\beq
    \label{eq:ranging}
    y = \f{c}{2} \left( (\mathtt{R}^2 - \mathtt{T}^1) - \f{(\mathtt{R^3} - \mathtt{R^2})}{(\mathtt{T^3} - \mathtt{T^2})} \left( \mathtt{T}^2 - \mathtt{R}^1 \right) \right),
\eeq
where $c$ is the speed of light.

When two UWB transceivers are ranging with one another, all other transceivers are passively listening-in on the messages transmitted by the ranging transceivers as in \cite{Shalaby2024passive}. Transceiver $\ell$ in Figure \ref{fig:ranging_protocol} performs this role, where it passively listens-in on all message transmissions, as shown by the blue arrows, and timestamps the $k^\text{th}$ message as $\mathtt{P}^k$.

Transceivers $i$, $j$, and $\ell$ \jln{all} need to know the DS-TWR timestamps recorded by transceivers $i$ and $j$ to be able to compute the range measurement \eqref{eq:ranging}. Transceiver $j$ embeds the timestamps $\mathtt{R}^1$, $\mathtt{T}^2$, and $\mathtt{T}^3$ in the last message it transmits, and thus these timestamps are available at transceivers $i$ and $\ell$. To do so, the timestamp $\mathtt{T}^3$ is predetermined in advance by the microprocessor, and the transmission occurs when Transceiver $j$'s clock reads $\mathtt{T}^3$. Meanwhile, transceiver $i$ transmits a fourth message that is not timestamped, shown in Figure \ref{fig:ranging_protocol} in red, with the sole purpose of broadcasting the timestamps $\mathtt{T}^1$, $\mathtt{R}^2$, and $\mathtt{R}^3$ to the other transceivers so they can all keep track of the timestamps and compute the range measurement.

\subsubsection{Media-Access Control}

The term media-access control (MAC) \jln{refers to} algorithms that control 
when each pair of transceivers 
\jln{initiates a ranging transaction,}
while preventing other transceivers from ranging during this time window to \jln{avoid} message collisions. 
The MAC algorithm used here involves pre-defining a sequence of unique ranging pairs, which is known to all robots. Each robot keeps track of which pair in the sequence is currently ranging by passively listening-in on neighbouring ranging transceivers, and then initiates a TWR transaction when it is its turn to do so. The sequence of ranging pairs is repeated when all pairs have successfully completed ranging transactions.

Note that the sequence of pairs does not contain ranging instances between two transceivers on the same robots, and only transceivers on robots initiate TWR transactions, as fixed transceivers, or \emph{anchors}, are not connected to a computer that can command them to initiate a ranging transaction. This also means that, in the dataset, UWB data is only available as recorded by the quadcopters, and information available only at the anchors, such as passive-listening measurements recorded by the anchors, are not made available.

\subsubsection{Message-Quality Metrics}

\begin{figure}
    \centering
    \includegraphics[width=\columnwidth]{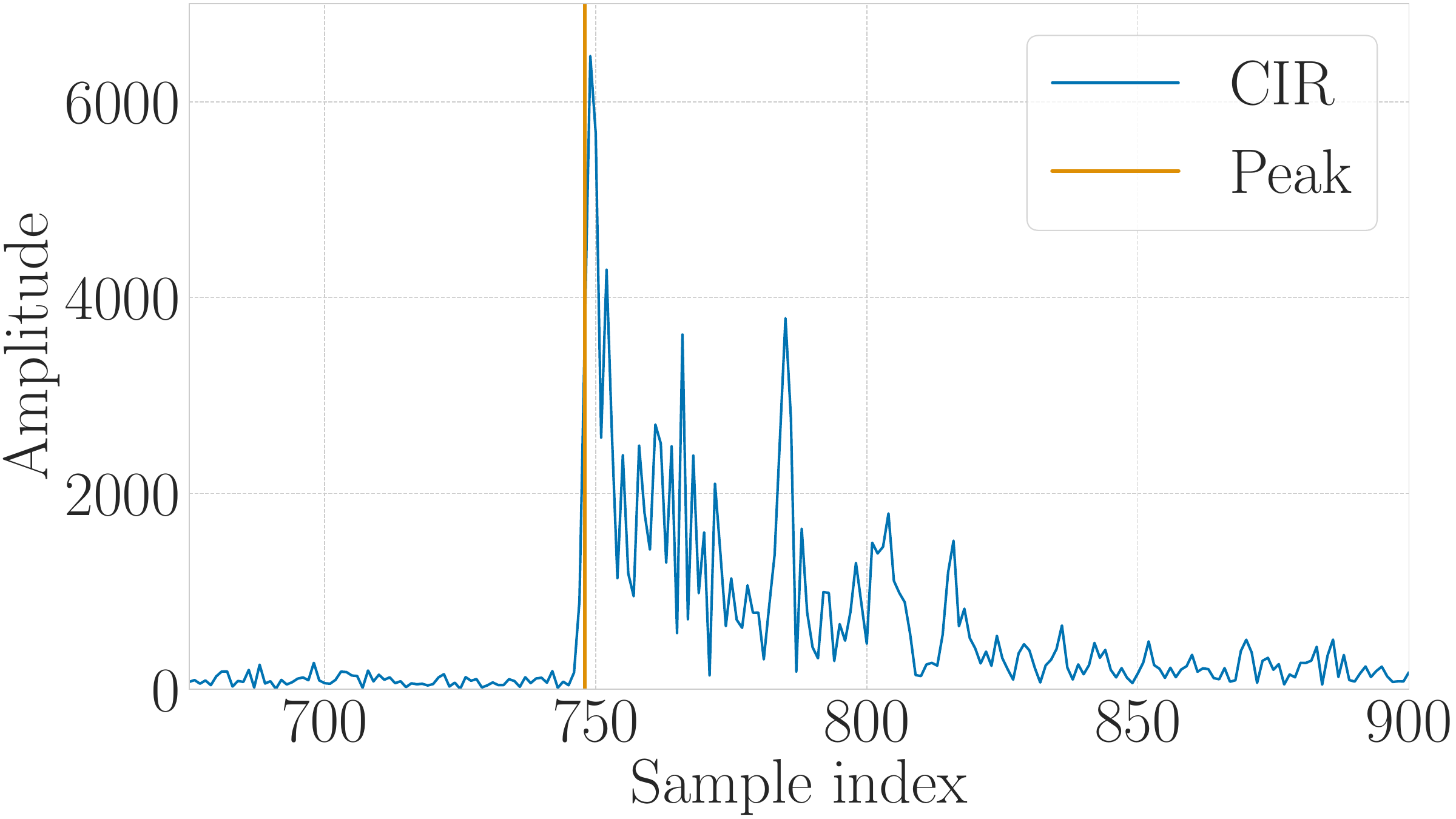}
    \caption{An example of a CIR reading for one ranging transaction, where the orange vertical line indicates the LDE's computed peak.}
    \label{fig:cir_example}
\end{figure}

In addition to the timestamps, various other metrics are recorded during UWB ranging. 
The CIR is a particularly rich source of information, representing the evolution of 
\jln{the signal recorded at a receiver} from a short pulse emitted by 
\jln{a transmitter}.
Onboard the DWM1000 module is a leading-edge detection (LDE) algorithm that attempts 
to find the \jln{first} edge in the CIR, which is then used to timestamp reception. 
The first peak is assumed to correspond to the \jln{signal following the} direct path 
between the two communicating transceivers, \jln{whereas} subsequent peaks 
\jln{may originate from} reflections from objects in the environment. The output of 
the onboard LDE is also recorded. \ijrr{Both the CIR and the index of the first edge are collected using the DW1000 device API\footnote{\url{https://github.com/lab11/dw1000-driver/tree/master}} \cite[Sections~5.59-5.60]{decawave_dw1000_api_guide}}, and an example of CIR data and the LDE's detected peak 
is shown in Figure~\ref{fig:cir_example} for one ranging transaction in the dataset.

The received signal power of the ``first-path signal'', as reported by the DWM1000 module, is also recorded. Lastly, clock-skew measurements generated using carrier-frequency offset estimation \citep{doltic2018range} are also recorded as part of this dataset to provide insight into the relative rates of the clocks of the different transceivers, and can be used for clock-state estimation (see, e.g., \cite{Cano2019}). 

The CIR, received signal power, and skew measurements are only made available for the first and third message reception of a DS-TWR instance. This is due to the processing time required to extract these metrics that might result in the transceiver missing the last reception. Additionally, it is expected that the second and third messages share similar properties. 
Lastly, when the CIR is collected, 
\jln{the rate of ranging measurements}
is reduced significantly as each CIR \jln{measurement} takes 
\jln{time} to be read from the registers and transferred to the computer for storage. 
As such, some experiments \jln{perform ranging at a lower rate to}
collect CIR data, 
while other experiments perform ranging at a higher rate \jln{but
without collecting CIR data.}

\section{Sensor Calibration} \label{sec:sensor_calibration}
\subsection{Inertial Measurement Unit} \label{subsec:imu_calib}


Each quadcopter has two IMUs, one connected to the Pixhawk autopilot computer, 
and one embedded in the Realsense D435i camera. Each IMU consists of a 3-axis gyroscope, 
measuring the angular velocity, and a 3-axis accelerometer measuring the specific force, 
both resolved in the IMU's own reference frame. 

The IMU measurement model assumed in this dataset 
\jln{includes}
two sources of error, an additive white noise and a slowly-varying sensor bias \cite[Ch. 9.4]{Barfoot2024}. As such, for an IMU reference frame $\mc{F}_i$, the gyroscope measurement is modelled as 
\beq
     \mbf{u}_i^{\text{gyr}}(t) = \mbs{\omega}_i^{ia}(t) - \mbs{\beta}_i^\text{gyr}(t) - \mbf{w}_i^\text{gyr}(t),
\eeq
where $\mbs{\omega}^{ia} \in \mathbb{R}^3$ is the angular velocity of $\mc{F}_i$ relative to $\mc{F}_a$, $\mbs{\beta}^\text{gyr}$ is the gyroscope bias, and $\mbf{w}^\text{gyr}$ is continuous-time white Gaussian gyroscope process noise. Meanwhile, the accelerometer measurement is modelled as 
\beq
     \mbf{u}_i^{\text{acc}}(t) = \mbf{f}_i(t) - \mbs{\beta}_i^\text{acc}(t) - \mbf{w}_i^\text{acc}(t),
\eeq
where $\mbf{f}$ is the specific force, $\mbs{\beta}^\text{acc}$ is the accelerometer bias, and $\mbf{w}^\text{acc}$ is continuous-time white Gaussian accelerometer process noise.

Letting $w$ denote any of the gyroscope and accelerometer process noise components, this single-channel process noise is modelled as Gaussian white noise with $\mathbb{E} [w (t)] = 0$ and
\begin{align*}
    \mathbb{E} [w(t_1) w(t_2)] = \sigma_\text{n}^2 \delta (t_1 - t_2),
\end{align*}
where $\mathbb{E} [\cdot]$ denotes the expectation operator. Here, $\delta(\cdot)$ denotes \emph{Dirac's delta function}, and $\sigma_\text{n}^2$ is the process-noise \emph{power spectral density} (PSD).

Meanwhile, letting $\beta$ denote any of the gyroscope and accelerometer bias components, 
this single-channel bias is modelled as the combination of two terms \cite[Ch. 2.5]{Pupo2016CharacterizationOE}: 
\begin{itemize}
    \item \emph{bias instability}, a high-frequency autocorrelated noise, and 
    \item \emph{random walk}, a low-frequency autocorrelated noise.
\end{itemize}
Therefore, the bias process model is modelled mathematically as 
\begin{align}
    \dot{\beta} (t) = -\f{1}{\tau} \beta (t) + w^\text{b} (t) + w^\text{r} (t), \label{eq:bias_process}
\end{align}
where $\tau$ is known as the \emph{correlation time}, $w^\text{b}$ is Gaussian white noise with $\mathbb{E}[w^\text{b}(t)] = 0$, 
\begin{align*}
    \mathbb{E} [w^\text{b}(t_1) w^\text{b}(t_2)] = \f{2}{\tau} \sigma_\text{b}^2 \delta(t_1-t_2),
\end{align*}
and $w^\text{r}$ is Gaussian white noise with $\mathbb{E}[w^\text{r}(t)] = 0$, 
\begin{align*}
    \mathbb{E} [w^\text{r}(t_1) w^\text{r}(t_2)] = \sigma_\text{r}^2 \delta(t_1-t_2).
\end{align*}
The first two components in \eqref{eq:bias_process} correspond to the bias instability Gauss-Markov process, while the last component is due to the random walk. Typically, it is assumed that the correlation time is infinity, hence reducing \eqref{eq:bias_process} to just
\begin{align*}
    \dot{\beta} (t) = w^\text{r} (t);
\end{align*}
nonetheless, all the quantities to allow the usage of the higher-fidelity model \eqref{eq:bias_process} are reported in this dataset. 

\begin{figure}[h]
    \centering
    \includegraphics[width=\columnwidth]{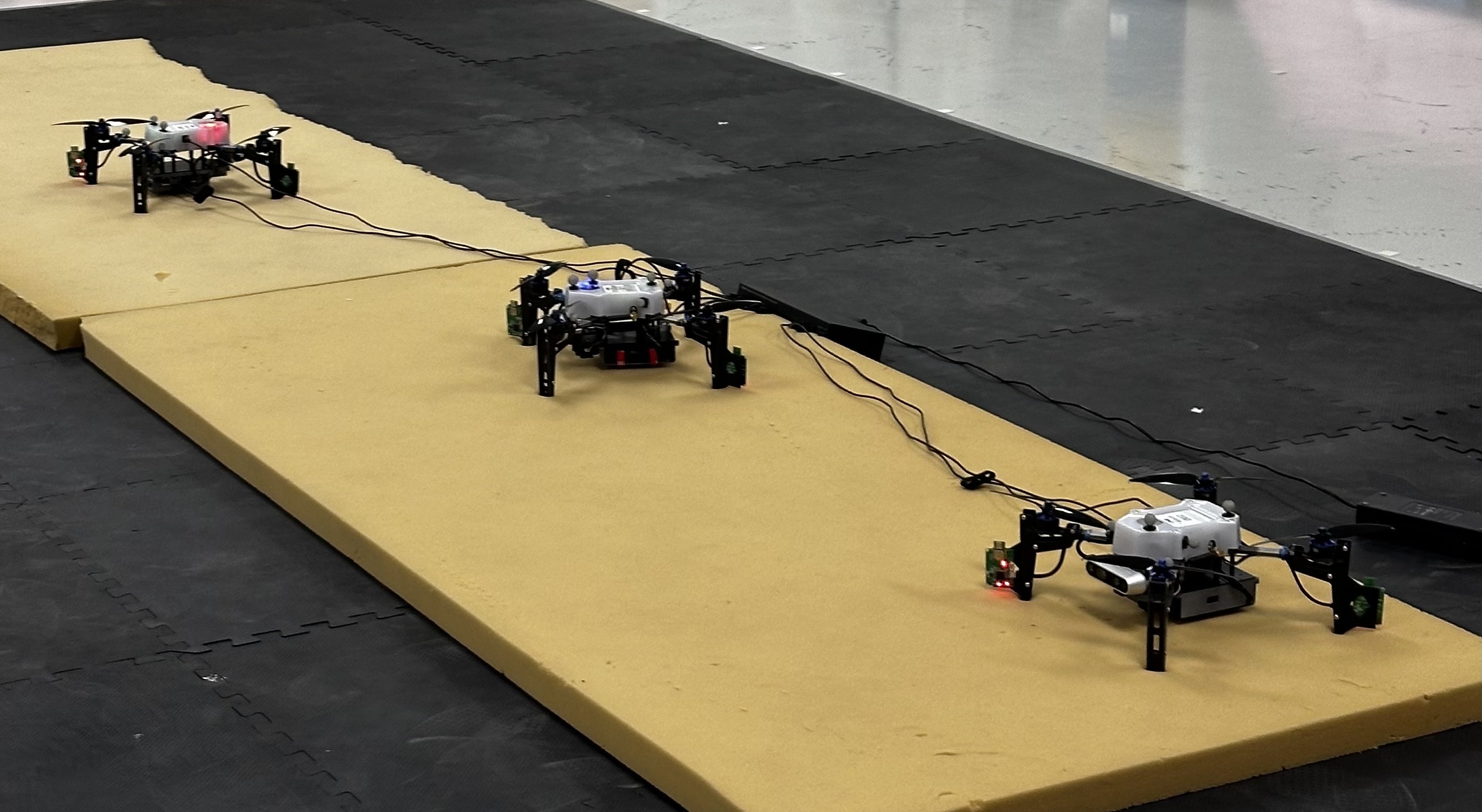}
    \caption{The three quadcopters, placed on foam and connected to an external power source, during the 16-hour-long static experiment to calibrate the onboard IMUs.}
    \label{fig:allan}
\end{figure}

To extract $\sigma_\text{n}$, $\tau$, $\sigma_\text{b}$, and $\sigma_\text{r}$, the quadcopters are placed on foam, as shown in Figure \ref{fig:allan}, connected to an external power source, and data from the IMU measurements are collected for 16 hours while the robots remain static. UWB data, although unnecessary for the IMU calibration, are also recorded. All this data is provided in the dataset in the \texttt{sensor\_calibration/imu\_calib} directory. The \texttt{allan\_variance\_ros}\footnote{\url{https://github.com/ori-drs/allan_variance_ros}} package is then used to perform an Allan variance analysis \citep{Pupo2016CharacterizationOE}, and the computed IMU noise and bias characteristics are reported in Tables \ref{tab:imu_noise} and \ref{tab:imu_bias}, respectively. \ijrr{The calibration results for the different IMUs on the three quadcopters are also provided in the development kit in the \texttt{config/imu} directory.}

\ijrr{In addition, the development kit provides an estimate of the IMU biases over time for all experiments by utilizing the motion capture data. This is done by computing a ``ground-truth'' IMU measurement from the ground-truth pose information, and then computing the difference between the recorded IMU measurements and the ground-truth IMU measurements. However, as these values would be corrupted by sensor noise, the generated IMU bias estimates are smoothed using spline fitting to reduce the effect of white noise. These smoothed IMU bias estimates are not part of the dataset but are generated using the development kit, and an example of how to access them is provided in \texttt{examples/visualize\_imu.py}. This data can then be used for evaluating algorithms that attempt to estimate IMU biases, or to remove IMU biases from the IMU data for early-stage research and development.}

\begin{table}[h]
\small\sf\centering
\caption{The IMU noise characteristics extracted using an Allan-variance analysis on 16-hour-long static experiments.}
\begin{tabular}{cccc}
& & \textbf{Accelerometer} & \textbf{Gyroscope} \\
\cmidrule(lr){3-3}\cmidrule(lr){4-4}
IMU & Axis & Random Walk & Random Walk  \\
& & $m/s/\sqrt{s}$ & $^\circ / \sqrt{s}$ \\
\midrule
\multirow{3}{*}{ifo001 camera}& x & 0.00124 & 0.00596  \\
& y & 0.00174 & 0.00694 \\
& z & 0.00182 & 0.00546\\
\midrule
\multirow{3}{*}{ifo001 PX4}& x & 0.00164 & 0.00799  \\
& y & 0.00173 & 0.00723 \\
& z & 0.00240 & 0.03126 \\
\midrule
\multirow{3}{*}{ifo002 camera}& x & 0.00112 & 0.00588 \\
& y & 0.00158 & 0.00787 \\
& z & 0.00219 & 0.00525 \\
\midrule
\multirow{3}{*}{ifo002 PX4}& x & 0.00280 & 0.00732 \\
& y & 0.00320 & 0.00848 \\
& z & 0.00627 & 0.02171 \\
\midrule
\multirow{3}{*}{ifo003 camera}& x & 0.00130 & 0.00734 \\
& y & 0.00151 & 0.00836 \\
& z & 0.00174 & 0.00582 \\
\midrule
\multirow{3}{*}{ifo003 PX4}& x & 0.00331 & 0.00812 \\
& y & 0.00437 & 0.00845 \\
& z & 0.00589 & 0.01397 \\
\end{tabular}
\label{tab:imu_noise}
\end{table}

\begin{table*}[h]
\footnotesize\sf\centering
\caption{The IMU bias characteristics extracted using an Allan-variance analysis on 16-hour-long static experiments.}
\begin{tabular}{cccccccc}
& & \multicolumn{3}{c}{\textbf{Accelerometer Bias}} & \multicolumn{3}{c}{\textbf{Gyroscope Bias}} \\
\cmidrule(lr){3-5}\cmidrule(lr){6-8}
IMU & Axis & Correlation Time & Instability & Random Walk & Correlation Time & Instability & Random Walk \\
& & $s$ & $m/s^2$ & $m/s^2/\sqrt{s}$ & $s$ & $^\circ/s$ & $^\circ/s/\sqrt{s}$ \\
\midrule
\multirow{3}{*}{ifo001 camera}& x & 31.3 & 0.00032 & 0.00012 & 90.6 & 0.00079 & 0.00015  \\
& y & 25.9 & 0.00067 & 0.00009 & 743.5 & 0.00073 & 0.00008  \\
& z & 20.5 & 0.00062 & 0.00021 & 182.3 & 0.00073 & 0.00009  \\
\midrule
\multirow{3}{*}{ifo001 PX4}& x & 6.9 & 0.00072 & 0.00023 & 56.5 & 0.00202 & 0.00052  \\
& y & 94.6 & 0.00073 & 0.00014 & 70.9 & 0.00169 & 0.00036  \\
& z & 110.6 & 0.00083 & 0.00015 & 22.0 & 0.01050 & 0.00208  \\
\midrule
\multirow{3}{*}{ifo002 camera}& x & 26.7 & 0.00030 & 0.00012 & 131.2 & 0.00064 & 0.00009  \\
& y & 17.5 & 0.00061 & 0.00014 & 127.8 & 0.00100 & 0.00013  \\
& z & 343.4 & 0.00041 & 0.00005 & 671.7 & 0.00049 & 0.00005  \\
\midrule
\multirow{3}{*}{ifo002 PX4}& x & 2.0 & 0.00103 & 0.00185 & 43.3 & 0.00248 & 0.00044  \\
& y & 1.6 & 0.00117 & 0.00141 & 38.6 & 0.00309 & 0.00117  \\
& z & 1.3 & 0.00197 & 0.00108 & 2.4 & 0.00834 & 0.02572  \\
\midrule
\multirow{3}{*}{ifo003 camera}& x & 49.4 & 0.00025 & 0.00006 & 90.4 & 0.00085 & 0.00015  \\
& y & 27.3 & 0.00045 & 0.00015 & 60.0 & 0.00149 & 0.00028  \\
& z & 695.1 & 0.00045 & 0.00005 & 673.4 & 0.00044 & 0.00005  \\
\midrule
\multirow{3}{*}{ifo003 PX4}& x & 1.6 & 0.00117 & 0.00155 & 107.5 & 0.00199 & 0.00031  \\
& y & 1.0 & 0.00149 & 0.00128 & 229.8 & 0.00287 & 0.00043  \\
& z & 1.2 & 0.00203 & 0.00028 & 2.0 & 0.00560 & 0.00851  \\
\end{tabular}
\label{tab:imu_bias}
\end{table*}

\subsection{Camera}

The Intel Realsense D435i cameras are calibrated using the \texttt{Kalibr}\footnote{\url{https://github.com/ethz-asl/kalibr}} toolbox. This toolbox allows intrinsic and extrinsic calibration of the Realsense D435i stereo camera, as well as spatial and temporal calibration of the IMUs in the camera and Pixhawk computer with respect to the camera system. Throughout this calibration procedure, the noise and bias characteristics for the IMUs obtained in Section \ref{subsec:imu_calib} are used. Additionally, a pinhole camera projection model and a radial-tangential distortion model \citep{Kanalla2006CameraModels} are assumed.

\begin{figure}
    \centering
    \includegraphics[trim={2cm 1.7cm 1.6cm 1.2cm}, clip, width=\columnwidth]{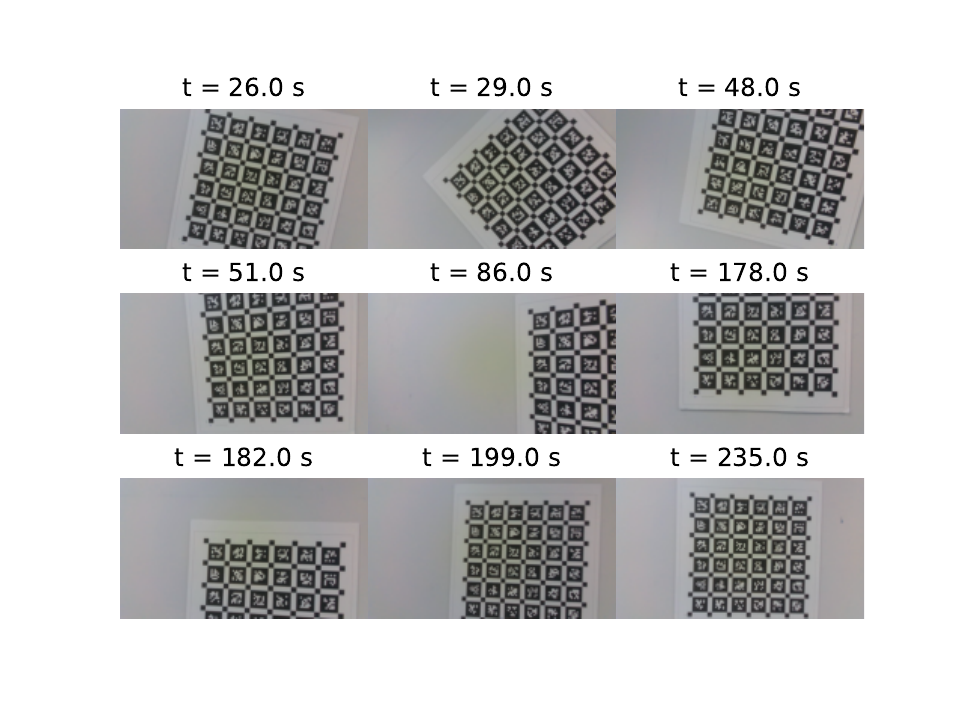}
    \caption{Samples of the images recorded by ifo001 during the camera-calibration procedure.}
    \label{fig:kalibr_calib}
\end{figure}

The calibration procedure is performed for one quadcopter at a time, by manually moving the robots randomly in front of a target consisting of a grid of AprilTags \citep{Olson2011} in a way that attempts to excite all axes of motion, for a span of $5$ \sh{minutes}. Some images taken by one of the quadcopters during the calibration procedure are shown in Figure \ref{fig:kalibr_calib}. The target used in this experiment contains six rows and six columns of AprilTags, each of size $6.25~\si{cm} \times 6.25~\si{cm}$ and with $1.875~\si{cm}$ separating each AprilTag. 

Throughout the calibration procedure, the quadcopters record the data published by each of the two cameras in the Realsense D435i, as well as the IMU measurements from both the IMU in the camera and the IMU connected to the Pixhawk computer. This allows calibrating the intrinsic parameters of each camera, such as the focal lengths, principal points, and the distortion coefficients. Additionally, the baseline between the two cameras is extracted, as well as the relative poses between the frames of both cameras and both IMUs for each quadcopter. Lastly, \texttt{Kalibr} also provides a time offset between the IMUs and the cameras. All the data and the standard output files of \texttt{Kalibr} are provided in the \texttt{sensor\_calibration/cam\_calib} directory\ijrr{, and a summary of the results is provided in the development kit in the \texttt{config/realsense} directory}.

\subsection{Ultra-wideband Transceivers}


The UWB measurements are corrupted by many sources of errors, even in LOS conditions. To improve the accuracy of the UWB measurements collected, the UWB calibration procedure presented in \cite{shalaby2023calib} is implemented. This involves solving for the delays in timestamping measurements at the antenna-level, and then finding a correlation between the received-signal power and the bias and standard deviation of the measurements. 

This procedure involves placing the UWB transceivers on mobile robots, and collecting range measurements in as many relative poses as possible between the robots. As such, the six UWB tags are first placed on the three quadcopters, which is the default setup for the majority of the experiments in this dataset. In the absence of anchors, two 4-minute-long experiments are collected with motion trajectories similar to that in \cite{shalaby2023calib}. Additionally, two more experiments are collected with similar motion, but with the six UWB tags replaced with the six UWB anchors instead to calibrate them as well. \ijrr{These experiments are available in the dataset in the \texttt{sensor\_calibration/uwb\_calib} directory.}

\begin{table}[h]
\small\sf\centering
\caption{The results of the antenna-delay calibration procedure for all 12 UWB transceivers.}
\begin{tabular}{c|cc}
& Tag ID & Antenna delay [ns] \\
\midrule
\multirow{6}{*}{Anchors}& 0 & -0.01825 \\
& 1 & -0.17987 \\
& 2 & -0.19663 \\
& 3 & -0.22567 \\
& 4 & -0.21826 \\
& 5 & -0.73767 \\
\midrule
\multirow{6}{*}{Tags} & 10 & -0.18471 \\
& 11 & -0.17734 \\
& 20 & -0.18259 \\
& 21 & -0.65886 \\
& 30 & -0.34798 \\
& 31  & -0.04120
\end{tabular}
\label{tab:antenna_delays}
\end{table}

\begin{figure}
    \centering
    \includegraphics[width=\columnwidth]{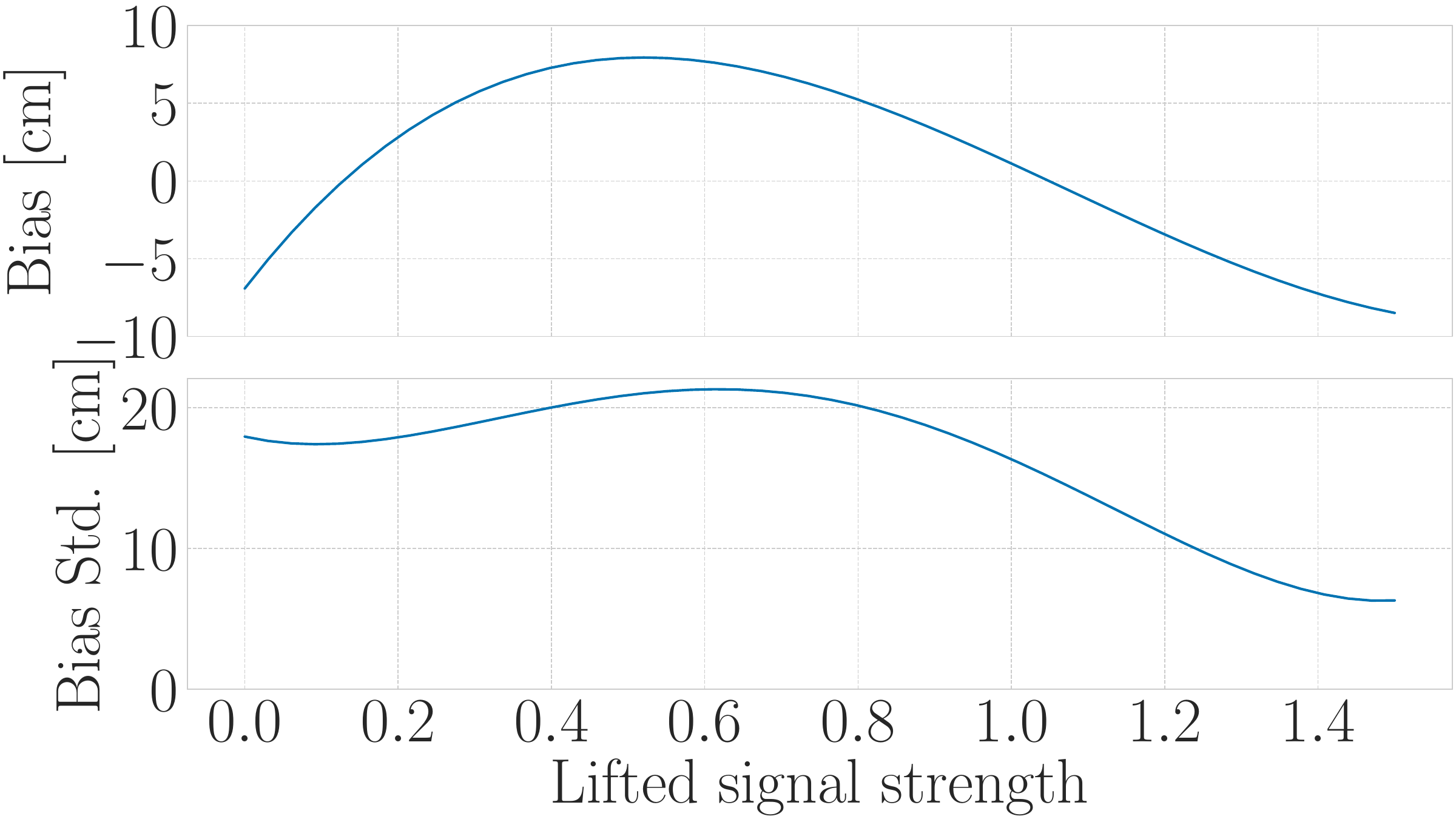}
    \caption{The fitted bias and standard deviation curves as a function of the lifted received signal strength, computed as per \cite{shalaby2023calib}.}
    \label{fig:calib_results}
\end{figure}

The \texttt{uwb\_calibration}\footnote{\url{https://github.com/decargroup/uwb_calibration}} package is used to extract the antenna delay estimates for the six UWB tags for each of the first two experiments, and for the six UWB anchors for each of the latter two experiments. A Huber loss \citep{Huber1964} is used for outlier rejection. The averaged delays for both experiments for each set of UWB transceivers are reported in Table \ref{tab:antenna_delays}. Additionally, the bias and standard deviation calibration as a function of the lifted received signal strength, calculated as per \cite{shalaby2023calib}, is computed for each of the four experiments, and their average is shown in Figure~\ref{fig:calib_results}. \ijrr{A summary of the calibration results is provided in the development kit, encoded into \texttt{config/uwb/uwb\_calib.pickle}.}

\begin{figure}
    \centering
    \includegraphics[width=\columnwidth]{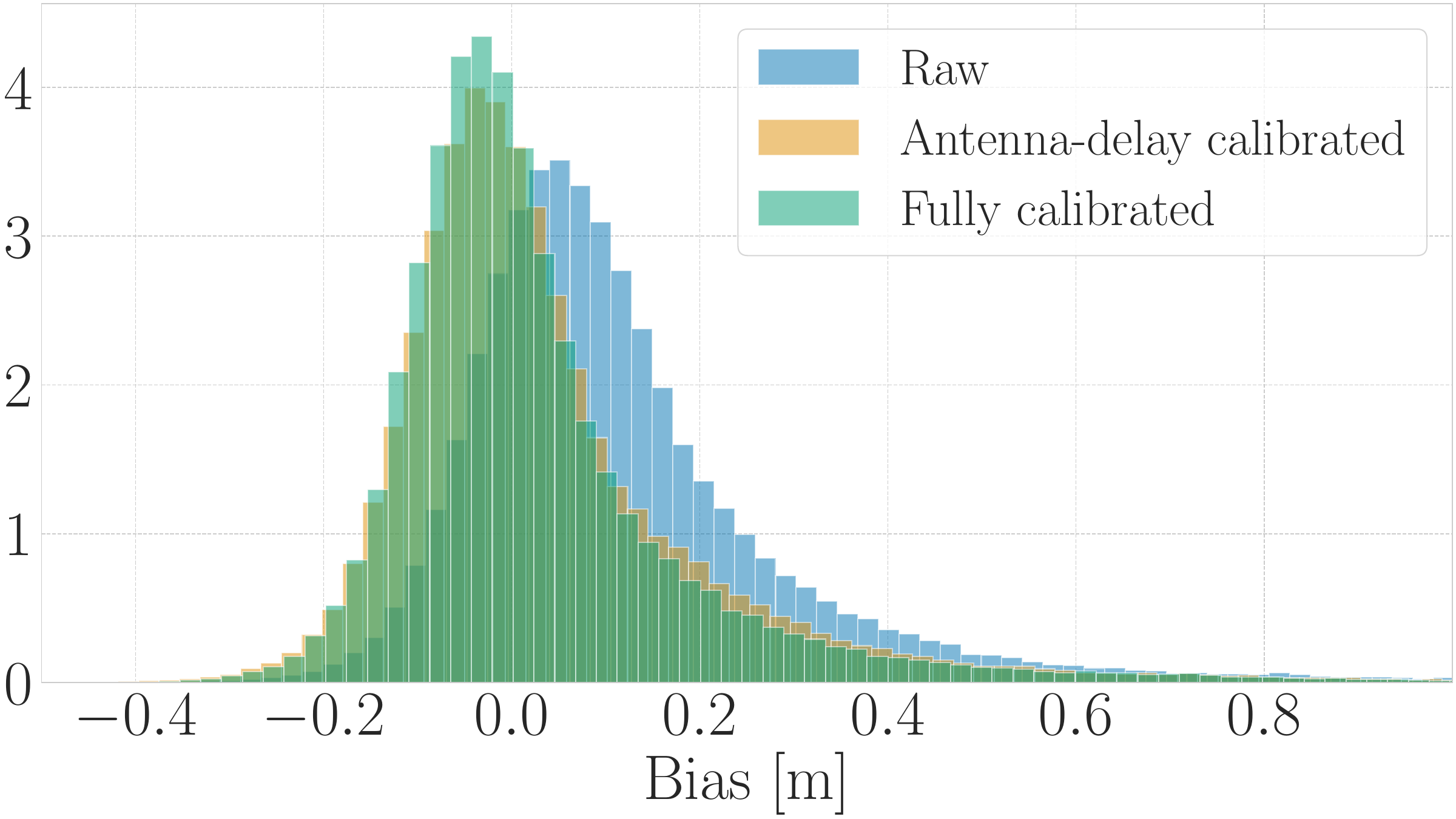}
    \caption{Histogram showing the bias distribution pre- and post-calibration, using a Huber loss.}
    \label{fig:bias_histogram}
\end{figure}

The reported antenna delay values and the averaged power curves are used to correct the range measurements. A histogram showing the bias distribution of the calibration data is shown in Figure~\ref{fig:bias_histogram}. Each experiment included in this dataset includes both raw and fully-calibrated range measurements and biases. 

\subsection{Time Synchronization}

Synchronization between the different computers' clocks is important to allow the robots to timestamp measurements from the different sensors in one common clock across the entire network.
At the beginning of each experiment, the clocks on the computers of all machines are manually synchronized using the \emph{Network Time Protocol} (NTP), which is intended to synchronize all clocks with Coordinated Universal Time (UTC) \citep{Mills2006NTP}. Specifically, the \texttt{Chrony}\footnote{\url{https://chrony-project.org/}} implementation is used, which claims to synchronize the computers' clocks to within sub-microsecond accuracy, but is hardware dependent. Note that the synchronized clocks are those of the computers, and the separate clocks on the UWB transceivers remain unsynchronized, which is addressed by the ranging protocol discussed in Section~\ref{subsubsec:ranging_protocol}.

\section{Data Collection} \label{sec:data_collection}
\begin{figure*}
\vspace{-0.5cm}
    \centering
    \subfloat[\centering Random~1]{{\includegraphics[width = 0.349\textwidth, trim={12cm 1cm 10cm 3cm}, clip]{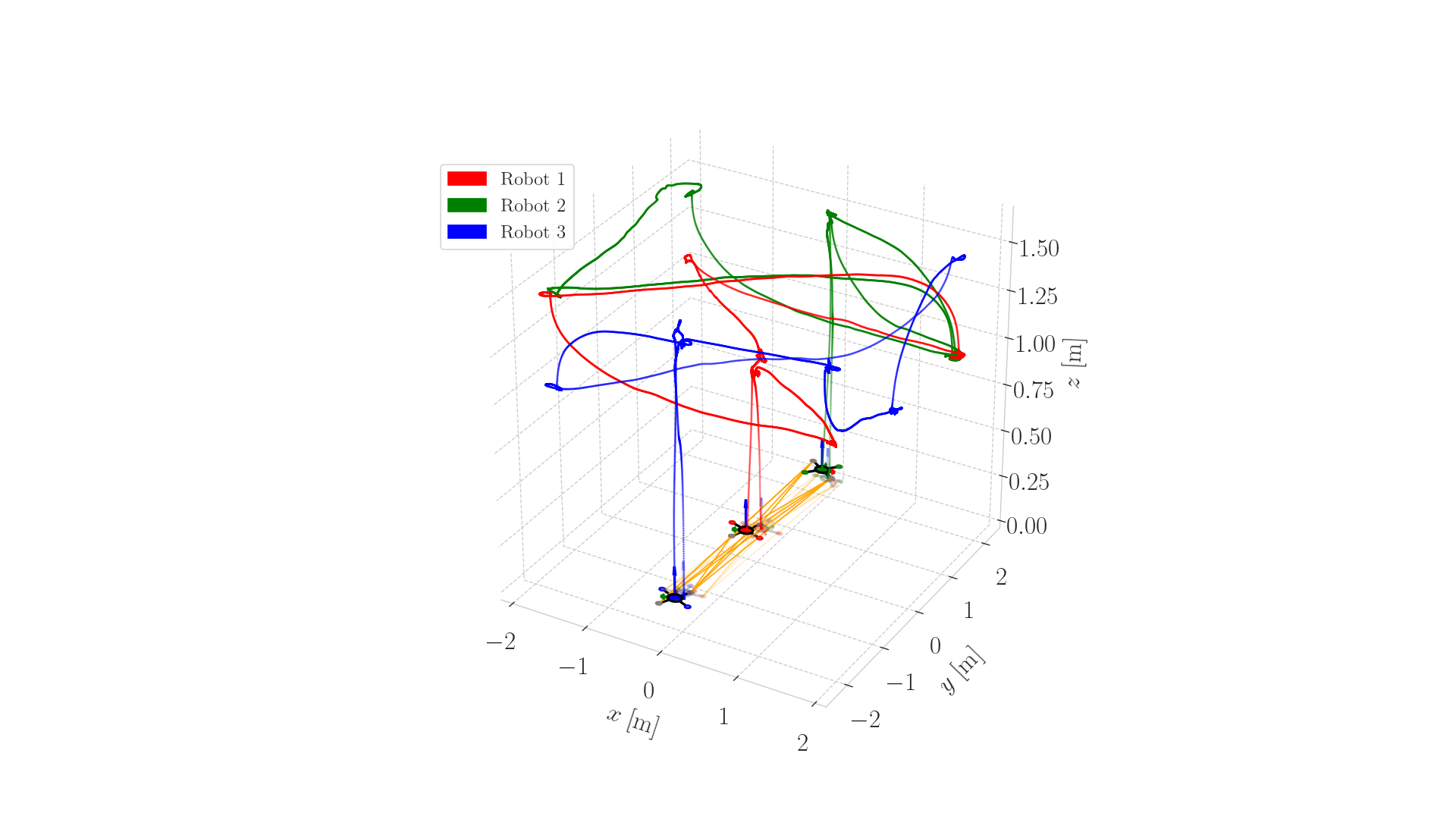}}}
    \subfloat[\centering Random~2]{{\includegraphics[width = 0.33\textwidth, trim={13cm 1cm 10cm 3cm}, clip]{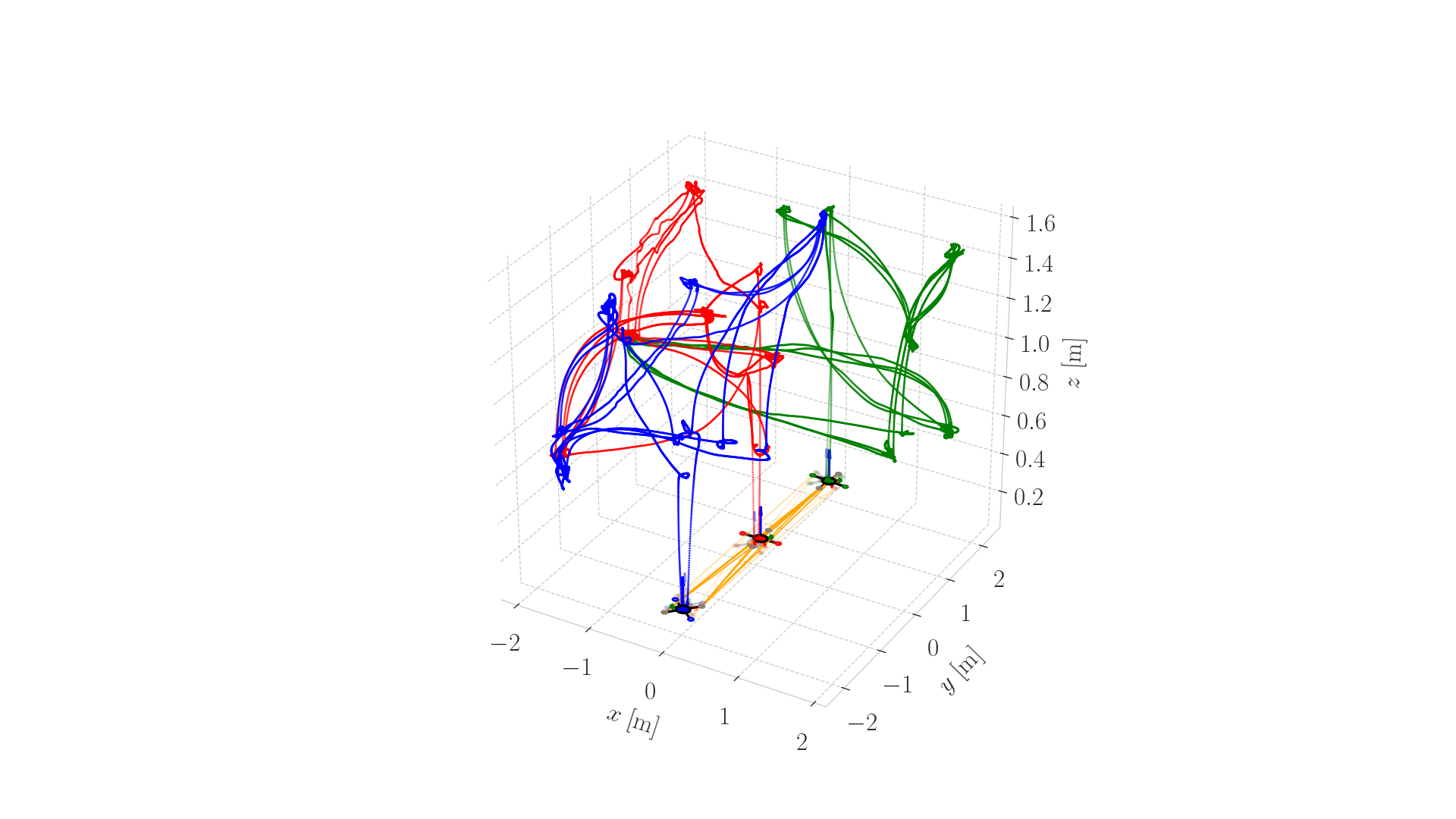}}}
    \subfloat[\centering Random~3]{{\includegraphics[width = 0.33\textwidth, trim={13cm 1cm 10cm 3cm}, clip]{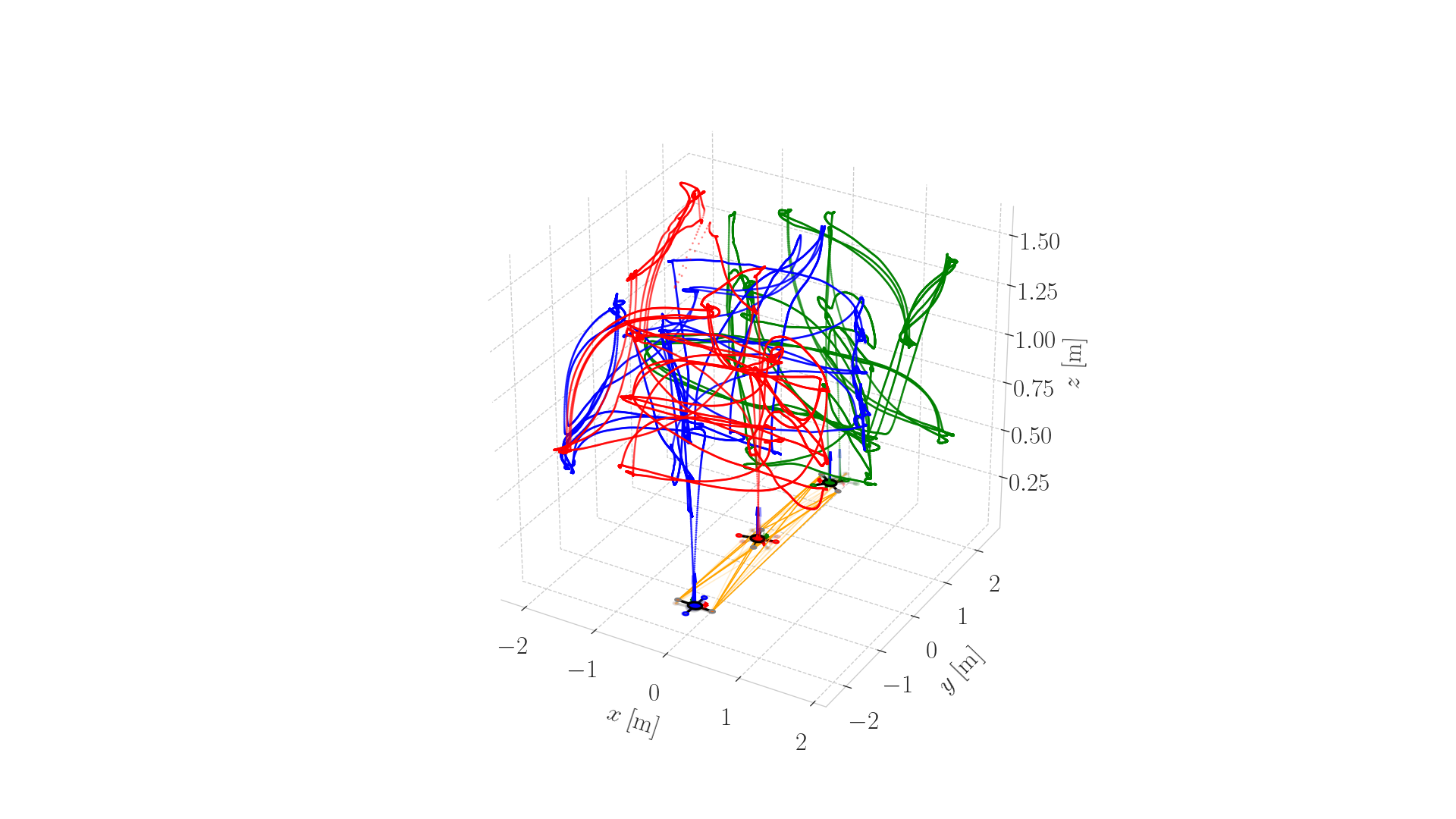}}}
    \quad
    \subfloat[\centering Moving Triangle]{{\includegraphics[width = 0.349\textwidth, trim={12cm 1cm 10cm 3cm}, clip]{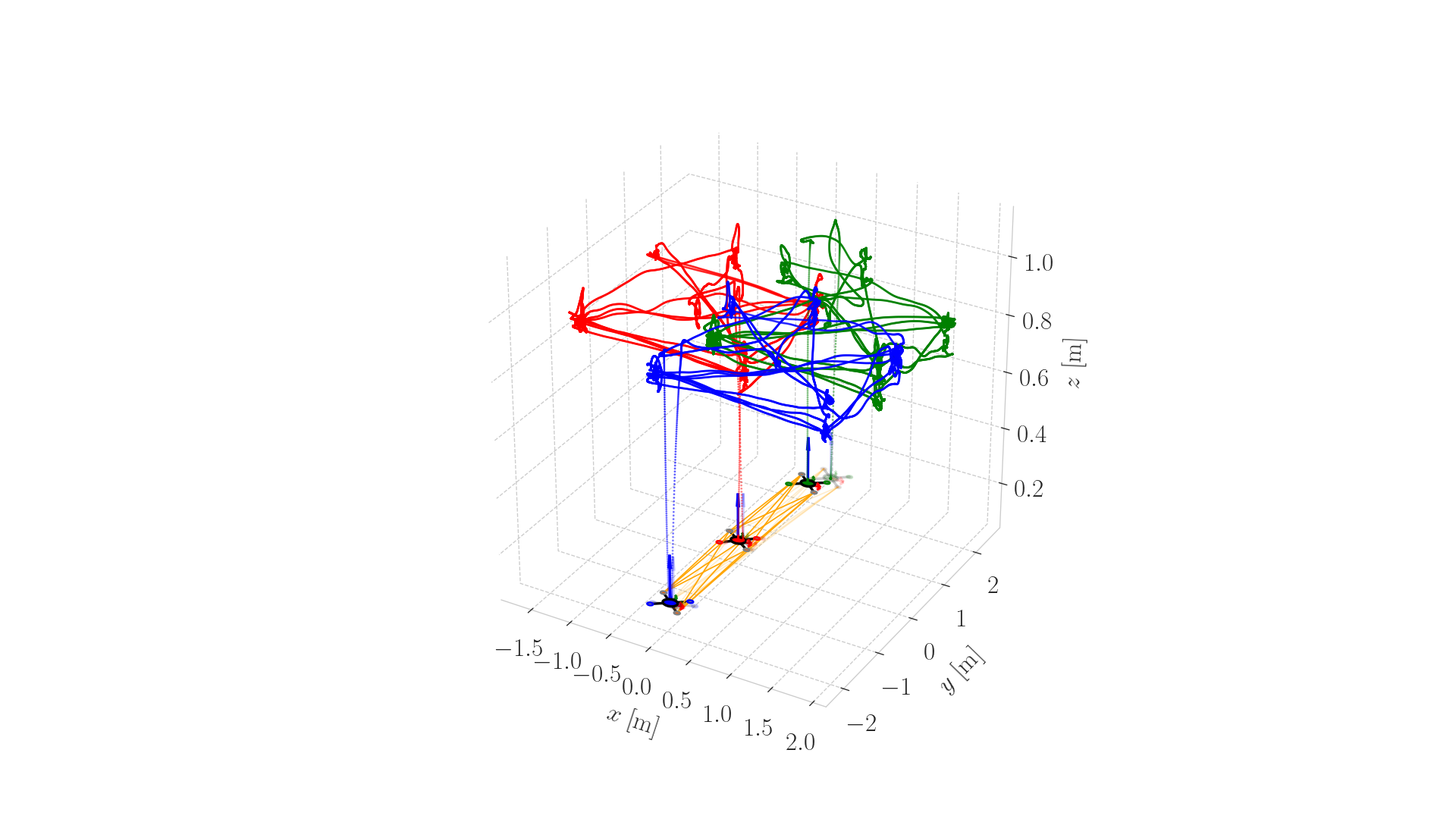}}}
    \subfloat[\centering Zigzag]{{\includegraphics[width = 0.33\textwidth, trim={13cm 1cm 10cm 3cm}, clip]{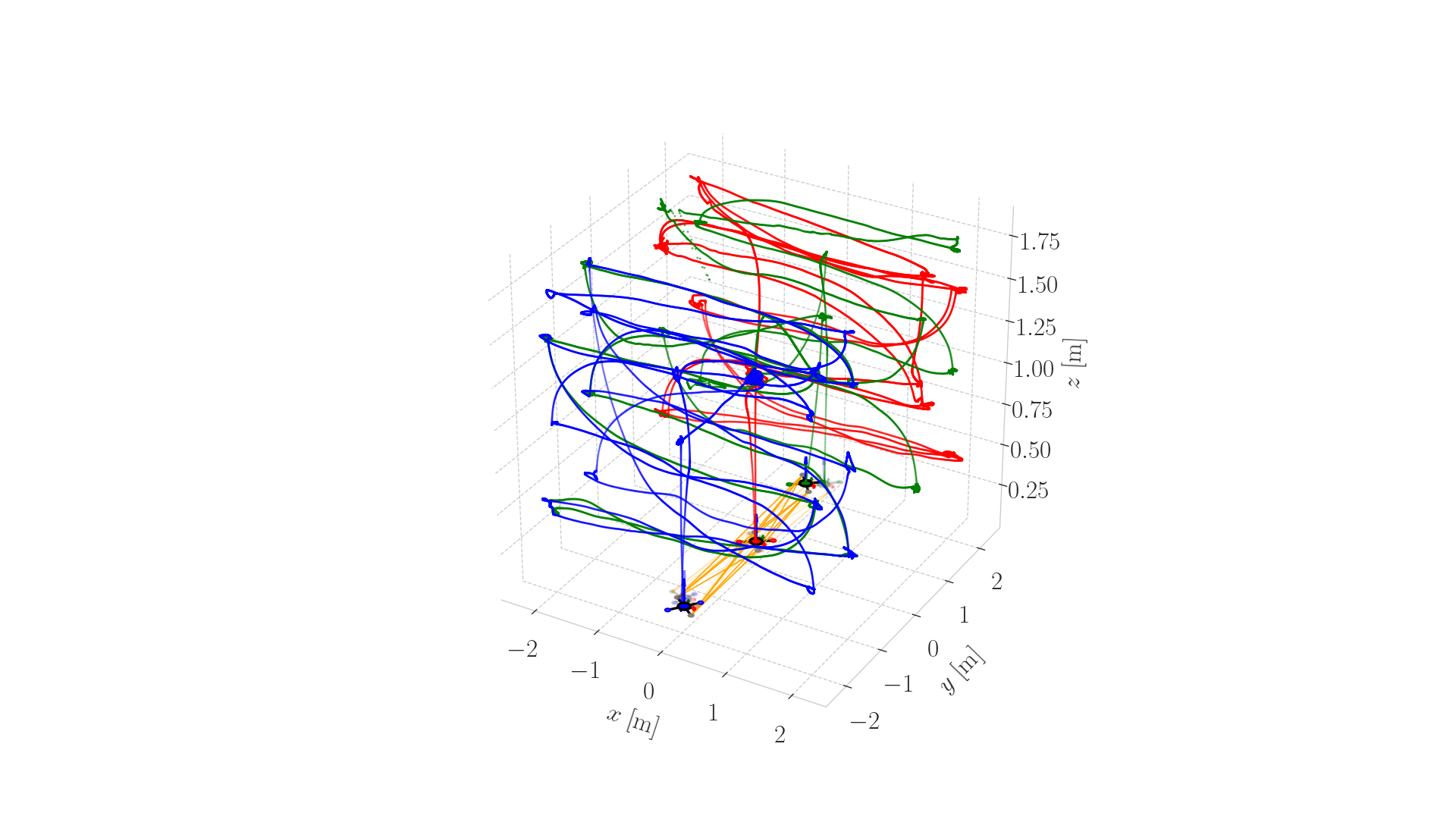}}}
    \subfloat[\centering Circular 2D]{{\includegraphics[width = 0.33\textwidth, trim={13cm 1cm 10cm 3cm}, clip]{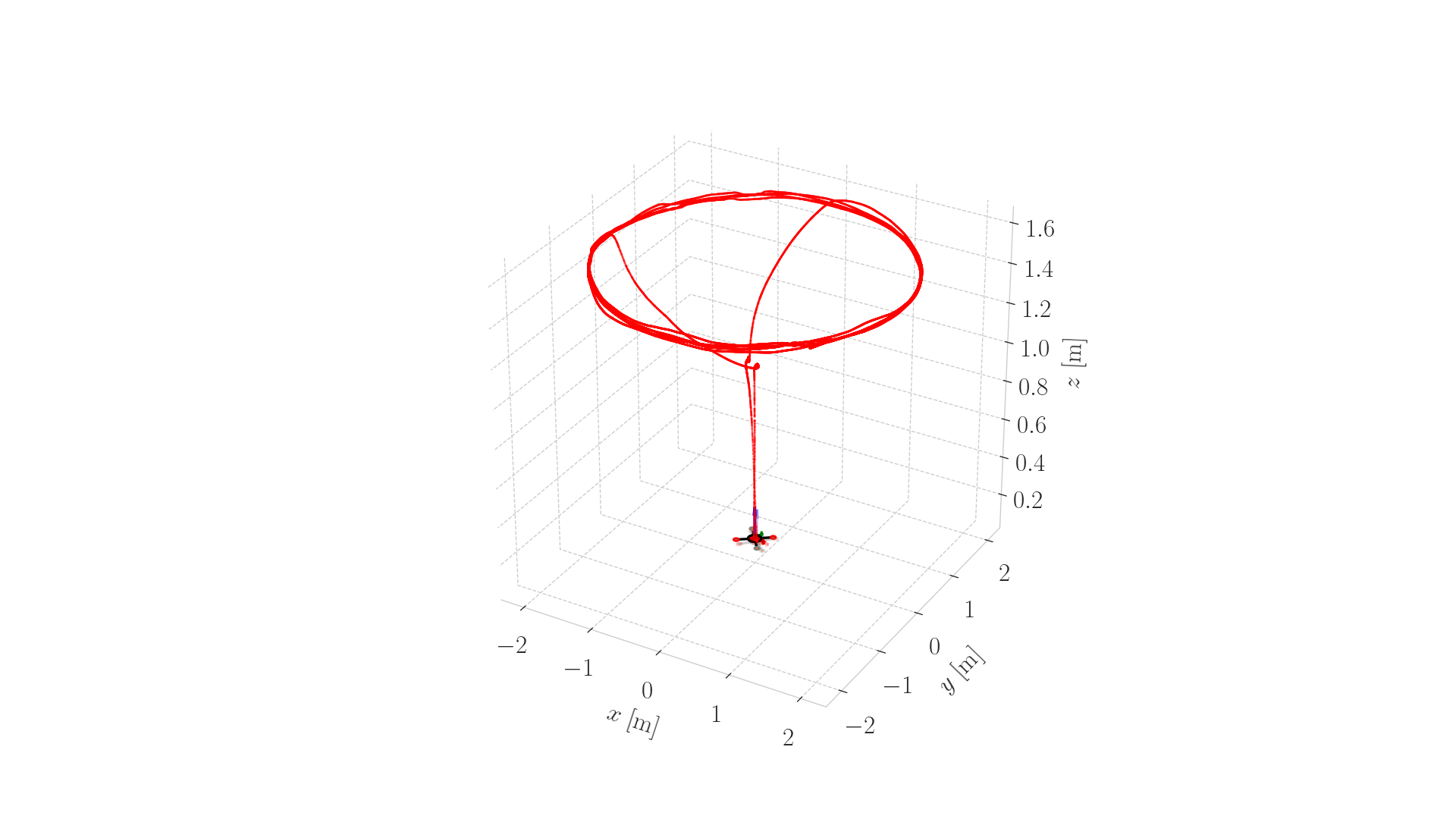}}}
    \quad
    \subfloat[\centering Circular 3D]{{\includegraphics[width = 0.349\textwidth, trim={12cm 1cm 10cm 3cm}, clip]{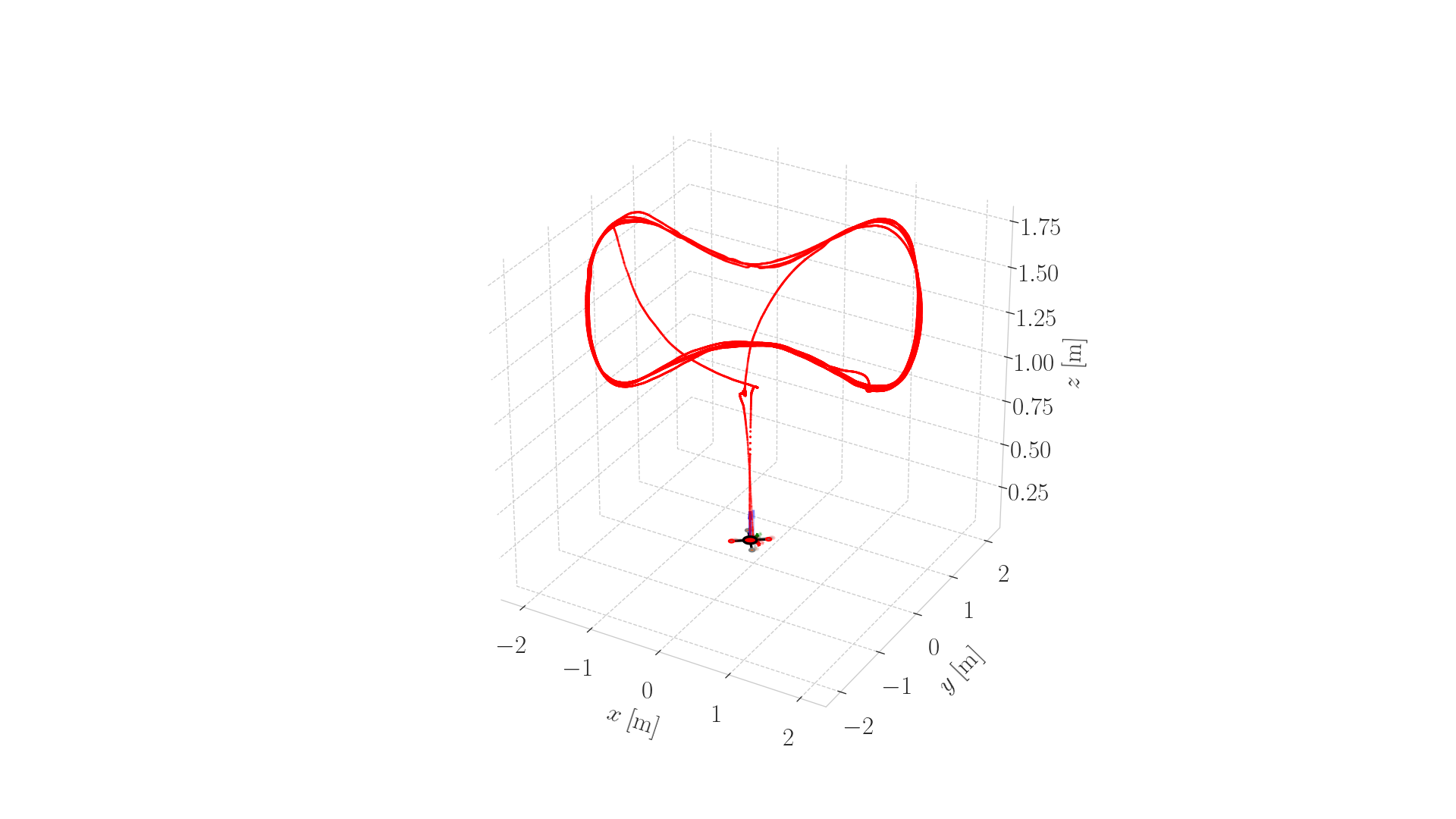}}}
    \subfloat[\centering Random]{{\includegraphics[width = 0.33\textwidth, trim={13cm 1cm 10cm 3cm}, clip]{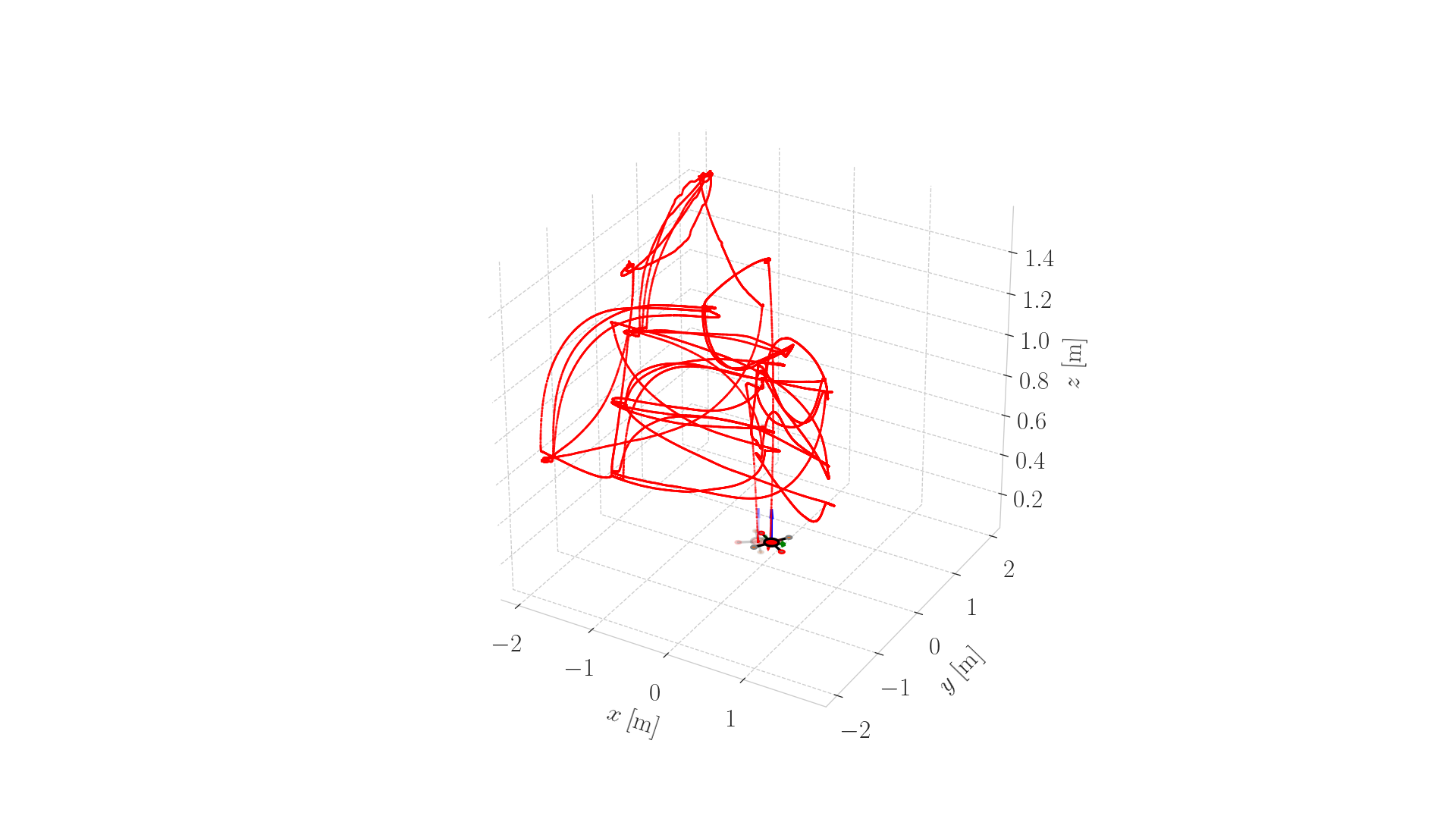}}}
    \subfloat[\centering Example of Remote Controlled (RC)]{{\includegraphics[width = 0.33\textwidth, trim={13cm 1cm 10cm 3cm}, clip]{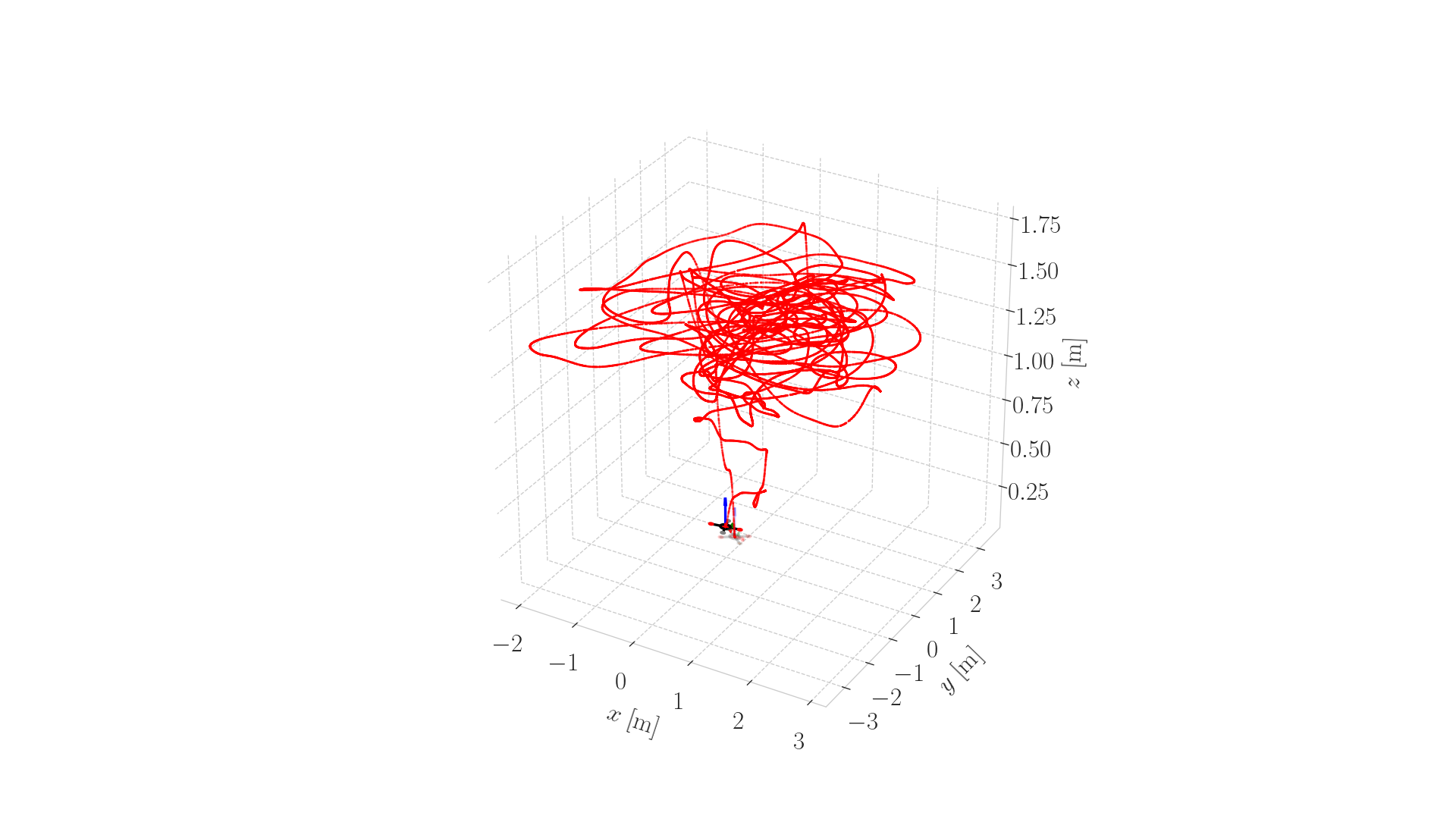}}}
    \quad
    \subfloat[\centering Linear]{{\includegraphics[width = 0.33\textwidth, trim={12cm 3cm 8cm 6cm}, clip]{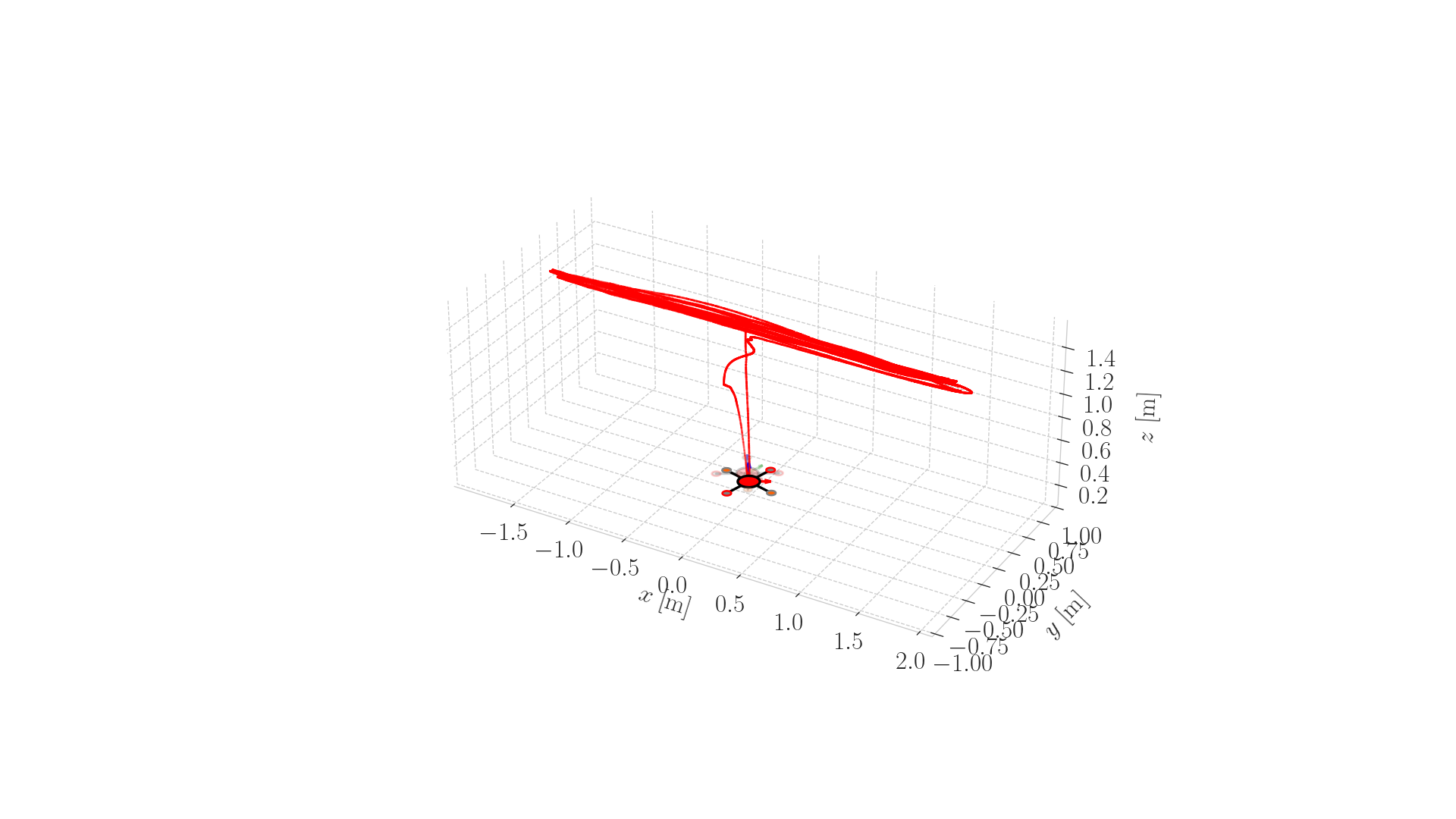}}}
    \caption{The 10 dynamic trajectories used in the collection of this dataset.
    }
    \label{fig:trajectories}
\end{figure*}

\begin{table*}
    \footnotesize\sf\centering
    \caption{\ijrr{A summary of the experiments performed in MILUV. The experiments are named as \texttt{scenario\_numberOfRobots\_motionTrajectory\_anchorConstellation}.}\vspace{8pt}}\label{tab:exp}
\begin{tabular}{
  c
  >{\centering\arraybackslash}p{1cm}
  >{\centering\arraybackslash}p{1cm}
  >{\centering\arraybackslash}p{1cm}
  >{\centering\arraybackslash}p{2cm}
  >{\centering\arraybackslash}p{1cm}
  >{\centering\arraybackslash}p{2cm}
  >{\centering\arraybackslash}p{1cm}
  c
}
    Experiment ID & No. of robots & No. of UWB tags/robot & Anchor Constellation & Trajectory & April Tags & Obstacles (wood, plastic, polystyrene) & CIR \\
    \midrule
    default\_3\_zigzag\_0 & 3 & 2 & 0 & Zigzag & \cmark & \xmark & \xmark \\
    \midrule
    default\_3\_zigzag\_1 & 3 & 2 & 1 & Zigzag & \cmark & \xmark & \xmark \\
    \midrule
    default\_3\_zigzag\_2 & 3 & 2 & 2 & Zigzag & \cmark & \xmark & \xmark \\
    \midrule
    default\_3\_random\_0 & 3 & 2 & 0 & Random~1 & \cmark & \xmark & \xmark\\
    \midrule
    default\_3\_random\_0b & 3 & 2 & 0 & Random~1 & \cmark & \xmark & \xmark \\
    \midrule
    default\_3\_random2\_0 & 3 & 2 & 0 & Random~2 & \cmark & \xmark & \xmark \\
    \midrule
    default\_3\_random3\_0b & 3 & 1 & 0 & Random~3 & \cmark & \xmark & \xmark \\
    \midrule
    default\_3\_random3\_1 & 3 & 2 & 1 & Random~3 & \cmark & \xmark & \xmark \\
    \midrule
    default\_3\_random3\_2 & 3 & 2 & 2 & Random~3 & \cmark & \xmark & \xmark \\
    \midrule
    default\_3\_movingTriangle\_0b & 3 & 2 & 0 & Moving Triangle & \cmark & \xmark & \xmark\\
    \midrule
    oneTagPerRobot\_3\_random\_0 & 3 & 2 & 0 & Random~1 & \cmark & \xmark & \xmark \\
    \midrule
    noAprilTags\_3\_random3\_2b & 3 & 2 & 2 & Random~3 & \xmark & \xmark & \xmark \\
    \midrule
    cir\_3\_random3\_0 & 3 & 2 & 0 & Random~3 & \cmark & \xmark & \cmark  \\
    \midrule
    cir\_3\_random3\_1 & 3 & 2 & 1 & Random~3 & \cmark & \xmark & \cmark  \\
    \midrule
    cir\_3\_random3\_2 & 3 & 2 & 2 & Random~3 & \cmark & \xmark & \cmark \\
    \midrule
    cirObstacles\_3\_random\_0 & 3 & 2 & 0 & Random~1 & \cmark & \cmark & \cmark \\
    \midrule
    \midrule
    default\_1\_random3\_0 & 1 & 2 & 0 & Random~3 & \cmark & \xmark & \xmark \\
    \midrule
    default\_1\_random3\_1 & 1 & 2 & 1 & Random~3 & \cmark & \xmark & \xmark \\
    \midrule
    default\_1\_random3\_2 & 1 & 2 & 2 & Random~3 & \cmark & \xmark & \xmark \\
    \midrule
    default\_1\_circular2D\_0 & 1 & 2 & 0 & Circular 2D & \cmark & \xmark & \xmark\\
    \midrule
    default\_1\_circular3D\_0 & 1 & 2 & 0 & Circular 3D & \cmark & \xmark & \xmark \\
    \midrule
    default\_1\_remoteControlHighPace\_0 & 1 & 2 & 0 & RC (high-paced) & \cmark & \xmark & \xmark \\
    \midrule
    default\_1\_remoteControlHighPace\_1 & 1 & 2 & 1 & RC (high-paced) & \cmark & \xmark & \xmark \\
    \midrule
    default\_1\_remoteControlHighPace\_2 & 1 & 2 & 2 & RC (high-paced) & \cmark & \xmark & \xmark \\
    \midrule
    default\_1\_remoteControlLowPace\_0 & 1 & 2 & 0 & RC (low-paced) & \cmark & \xmark & \xmark \\
    \midrule
    default\_1\_remoteControlLowPace\_0\_v2 & 1 & 2 & 0 & RC (low-paced) & \cmark & \xmark & \xmark \\
    \midrule
    oneTagPerRobot\_1\_random3\_0 & 1 & 1 & 0 & Random~3 & \cmark & \xmark & \xmark \\
    \midrule
    noAprilTags\_1\_random3\_2b & 1 & 2 & 2 & Random~3 & \xmark & \xmark & \xmark \\
    \midrule
    noAprilTags\_1\_remoteControlHighPace\_2b & 1 & 2 & 2 & RC (high-paced) & \cmark & \xmark & \xmark \\
    \midrule
    obstacles\_1\_random3\_0b & 1 & 2 & 0 & Random~3 & \cmark & \cmark & \xmark \\
    \midrule
    cir\_1\_circular2D\_0 & 1 & 2 & 0 & Circular 2D & \cmark & \xmark & \cmark \\
    \midrule
    cirOneTag\_1\_static\_0 & 1 & 1 & 0 & Static & \cmark & \xmark & \cmark \\
    \midrule
    cirObstacles\_1\_random3\_0 & 1 & 2 & 0 & Random~3 & \cmark & \cmark & \cmark \\
    \midrule
    cirObstaclesOneTag\_1\_static\_0 & 1 & 1 & 0 & Static & \cmark & \cmark & \cmark \\
    \midrule
    cirOneTagOneAnchor\_1\_linear\_0 & 1 & 1 & 0 & Linear & \cmark & \xmark & \cmark \\
    \midrule
    cirOneTagOneAnchor\_1\_angular\_0 & 1 & 1 & 0 & Rotational & \cmark & \xmark & \cmark \\
    \bottomrule
    \\\\\\
    \end{tabular}
\end{table*} 

The \jln{collected data} is broadly divided into two parts. As detailed below,
one portion of the experiments includes three quadcopters, and the other includes only one of three quadcopters.

\subsection{Trajectories}
Three quadcopters are used to collect data for five different trajectories, which are named as follows: Random~1, Random~2, Random~3, Zigzag, and Moving Triangle. A single quadcopter is also used to collect data for eight different trajectories. These are identified as Circular~2D, Circular~3D, Random, Remote-controlled (high-paced), Remote-controlled (low-paced), Linear, Rotational, and Static. All the trajectories taken by the three quadcopters and some of the trajectories of the single quadcopter experiments are shown in Figure~\ref{fig:trajectories}. 

For the single-quadcopter case, the human-operated remote-controlled trajectories follow a random pattern, which is not pre-defined. There are no speed limitations on the remote-controlled high-paced trajectories and the quadcopter attained a maximum translational speed of $4.418\,\si{m/s}$ in some experiments. For the low-paced trajectories, the quadcopter is restricted to a maximum translational speed of $0.7\,\si{m/s}$. Additionally, note that the Rotational and Static cases are not plotted since there is no translational movement as the quadcopter is hovering with a constant angular velocity mid-air for the former and is stationary for the latter. A combination of these trajectories and environmental setups, such as different anchor constellations, forms the basis of the experiments conducted in this work.

\subsection{Experiments}
For the majority of the experiments, the parameters varied are the number of robots, number of UWB tags per robot, number of anchors, anchor constellation, the presence or absence of AprilTags, the presence or absence of obstacles disrupting the LOS between UWB tags, and robot trajectories. For all the experiments with obstacles, the wooden, acrylic plastic, and polystyrene obstacles mainly disrupt the LOS from three anchor tags with IDs 4,~3,~and~1, respectively. Additional experiments are recorded with CIR data as well. The anchor constellations and AprilTag formations used in the experiments are shown in Figure~\ref{fig:anchors} and Figure~\ref{fig:april_tags}, respectively. Table~\ref{tab:exp} summarizes the experimental setup for each experiment. The table divides the experiments into two broad categories, one where there are three robots, and the other one with only one robot. \ijrr{A summary of the experiments can also be found in the development kit at \texttt{config/experiments.csv}.}

\begin{figure}[t]
    \centering
\includegraphics[trim={2cm 18cm 3cm 2cm}, clip, width=\columnwidth]{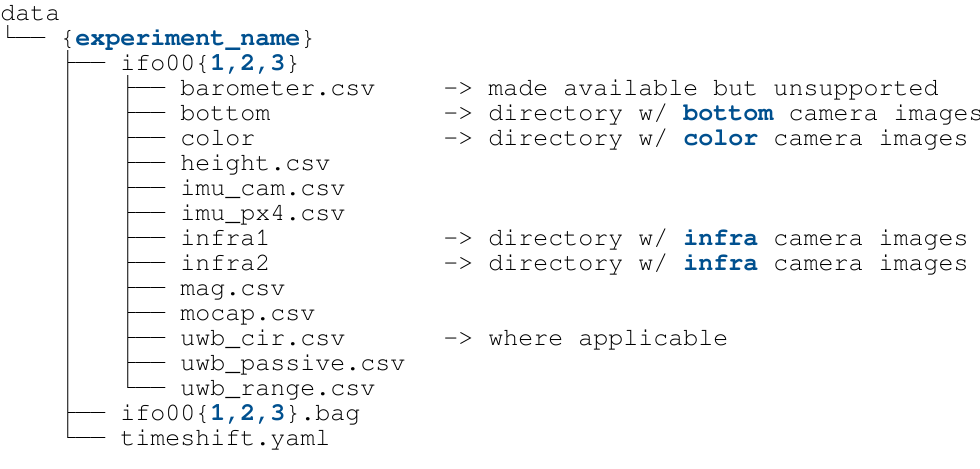}
    \caption{\ijrr{File structure of the MILUV dataset.}}
    \label{fig:miluv_tree}
\end{figure}



\section{Dataset Usage} \label{sec:dataset_usage}
\subsection{Data Format}

The data collected from each quadcopter's onboard sensors is provided as a collection of \texttt{csv} \jln{files}. 
The file structure of the dataset is shown in Figure~\ref{fig:miluv_tree}. For each experiment, a suite of sensors on each quadcopter provide measurements and their associated timestamps. 
The format of the \texttt{csv} files representing the data collected from each of the quadcopters \jln{is} shown in Table~\ref{tab:csvs}. Most of the headers are self-explanatory, except the headers in the UWB files, which are further elaborated in Table~\ref{tab:uwb_range}.

The data collected from the quadcopters' cameras is provided as a collection of \texttt{jpeg} files. For each experiment, a monocular downward-facing camera collects black-and-white images and the forward-facing D435i stereo camera collects monocular images in colour and infrared stereo images. The file name of each image corresponds to the timestamp, in seconds, at which it was collected, with an underscore acting as a decimal place.

\begin{table*}[t]
\footnotesize\sf\centering
\caption{Summary of the headers in the CSV files for each quadcopter ifo001, ifo002, and ifo003 in the dataset.}
\label{tab:csvs}
\begin{tabular}{|c|l|}
\hline
\textbf{File Name} & \textbf{CSV Columns}\\
\hline \hline
\texttt{height}& \texttt{timestamp, range} \\
\hline
\texttt{imu\_cam, imu\_px4}& 
\texttt{timestamp,	angular\_velocity.\{x,y,z\}, linear\_acceleration.\{x,y,z\}}\\
\hline
\texttt{mag}&\texttt{timestamp,	magnetic\_field.\{x,y,z\}}\\
\hline
\texttt{mocap} & \texttt{timestamp,	pose.position.\{x,y,z\}, pose.orientation.\{x,y,z,w\}}\\
\hline
\multirow{4}{*}{\texttt{uwb\_passive}}&\texttt{timestamp, timestamp\_n, my\_id, from\_id, to\_id, rx\{1,2,3\}, fpp\{1,2,3\},}\\
&\texttt{skew\{1,2,3\}, fpp\{1,2\}\_n, skew\{1,2\}\_n, fpp\{1,2\}\_n, skew\{1,2\}\_n,}\\
&\texttt{rx\{1,2,3\}\_raw, std, tx\{1,2,3\}\_n,	rx\{1,2,3\}\_n, range\_n, bias\_n,}\\
&\texttt{tx\{1,2,3\}\_raw\_n, rx\{1,2,3\}\_raw\_n, range\_raw\_n, bias\_raw\_n, gt\_range\_n}
\\
\hline
\multirow{3}{*}{\texttt{uwb\_range}}&
\texttt{timestamp, range, from\_id, to\_id, tx\{1,2,3\}, rx\{1,2,3\},}\\
&\texttt{fpp\{1,2\}, skew\{1,2\}, r\_iw\_a\_ifo00\{1,2,3\}, q\_ai\_ifo00\{1,2,3\}, gt\_range,}\\
&\texttt{bias, range\_raw, bias\_raw, tx\{1,2,3\}\_raw, rx\{1,2,3\}\_raw, std}
\\
\hline
\texttt{uwb\_cir}& \texttt{timestamp, my\_id, from\_id, to\_id, idx, cir}\\
\hline
\end{tabular}
\end{table*}

\begin{table*}[t]
\footnotesize\sf\centering
\caption{Description of the column headers in \texttt{uwb\_range.csv}, \texttt{uwb\_passive.csv}, and \texttt{uwb\_cir.csv} files.}
\label{tab:uwb_range}
\begin{tabular}{|c|l|}
\multicolumn{2}{c}{\texttt{uwb\_range.csv}}\\
\hline
\textbf{CSV Column Name} & \textbf{Description}\\
\hline \hline
\texttt{range}& The calibrated range measurement computed using (1). \\
\hline
\texttt{from\_id}& The ID of the initiating transceiver, Transceiver $i$ in Figure~\ref{fig:ranging_protocol}.\\
\hline
\texttt{to\_id}& The ID of the target transceiver, Transceiver $j$ in Figure~\ref{fig:ranging_protocol}.\\
\hline
\texttt{tx\{i\}} & The $i^{\text{th}}$ transmission in the ranging transaction, corresponding to $\mathtt{T}^i$ in Figure~\ref{fig:ranging_protocol}.\\
\hline
\texttt{rx\{i\}} & The $i^{\text{th}}$ reception in the ranging transaction, corresponding to $\mathtt{R}^i$ in Figure~\ref{fig:ranging_protocol}.\\
\hline
\texttt{fpp\{i\}} & The received signal power at the $i^{\text{th}}$ reception in the ranging transaction.\\
\hline
\texttt{skew\{i\}} &  DWM1000 skew measurement for the $i^{\text{th}}$ reception in the ranging transaction, computed using \cite{doltic2018range}.\\
\hline
\texttt{r\_iw\_a\_ifo00\{1,2,3\}} &  The position of \texttt{ifo00\{1,2,3\}} at the ranging transaction, in the Vicon frame, $\mc{F}_a$.\\
\hline
\texttt{q\_ai\_ifo00\{1,2,3\}} &  The quaternion of the \texttt{ifo00\{1,2,3\}} at the ranging transaction, in the Vicon frame, $\mc{F}_a$.\\
\hline
\texttt{gt\_range} & Ground truth range measurement.\\
\hline
\texttt{bias} &  The bias in the calibrated measurement, given by \texttt{range - gt\_range}.\\
\hline
\texttt{std} & The standard deviation of the measurement, computed using \cite{shalaby2023calib}.\\
\hline
\texttt{\{column\}\_raw} & \texttt{\_raw} represents the raw measurement before performing calibration using \cite{shalaby2023calib}.\\
\hline 
\multicolumn{2}{c}{}\\
\multicolumn{2}{c}{\texttt{uwb\_passive.csv}}\\
\multicolumn{2}{c}{*** The column headers in \texttt{uwb\_passive.csv} are the same as the ones given in the top table except the following given below ***}\\
\hline
\textbf{CSV Column Name} & \textbf{Description}\\
\hline \hline
\texttt{my\_id}& The ID of the passively-listening transceiver, Transceiver $l$ in Figure~\ref{fig:ranging_protocol}. \\
\hline
\texttt{rx\{i\}}& The $i^{\text{th}}$ passive listening measurement recorded in the ranging transaction, corresponding to $\mathtt{P}^i$ in 
Figure~\ref{fig:ranging_protocol}.\\
\hline
\multirow{2}{*}{\texttt{\{column\}\_n}}& Each passive listening measurement is matched to an entry in \texttt{uwb\_range.csv},\\
&and the \texttt{\_n} suffix indicates the matched data recorded from the neighboring ranging transceiver.\\
\hline
\multicolumn{2}{c}{}\\
\multicolumn{2}{c}{\texttt{uwb\_cir.csv}}\\
\multicolumn{2}{c}{*** Only table headers not described in the top and middle table are mentioned below ***}\\
\hline
\textbf{CSV Column Name} & \textbf{Description}\\
\hline \hline
\texttt{idx}& Index of the peak detected by DWM1000's onboard leading-edge detector (LDE). \\
\hline
\texttt{cir}& The accumulated channel impulse response.\\
\hline
\end{tabular}
\end{table*}

\subsection{Development Kit}

The MILUV development kit provided with this dataset, available at \url{https://github.com/decargroup/miluv}, contains Python scripts to help users get started using the MILUV dataset. It provides code to parse through the data in the form of a Python dictionary, keyed by robots and sensors. Other provided functions include keying sensor measurements and images by timestamps, which enables subsampling of the measurements. Sample code that makes use of these functions is shown in Figure~\ref{fig:devkit_example}. \ijrr{More detailed examples are provided in the development kit in the \texttt{examples} directory, for which documentation is available at \url{https://decargroup.github.io/miluv/}. The documentation website includes the file structure of the development kit, instructions on local or docker-based installation and getting started, tutorials for usage of the dataset and the development kit, and more.}

\begin{figure}
    \centering
    \begin{minipage}{\columnwidth}
\begin{lstlisting}[language=Python, numbers=none]
from miluv.data import DataLoader
import numpy as np

mv = DataLoader("default_3_random_0")

timestamps = np.arange(0, 10, 1)

# Fetching imu data for ifo001/ifo002
robots = ["ifo001", "ifo002"]  
sensors = ["imu_px4", "imu_cam"]  

data_at_timestamps = mv.data_from_timestamps(
    timestamps, robots, sensors
)
\end{lstlisting}
    \end{minipage}
    \caption{Example usage of the \lstinline[language=Python]{data_from_timestamps} function provided in the MILUV devkit}
    \label{fig:devkit_example}
\end{figure}


\subsection{Localization Benchmarking}

The provided data is benchmarked using three standard localization approaches: 
\begin{enumerate}
    \item visual-inertial odometry, with and without loop-closures, 
    \item UWB-inertial localization using an EKF, and
    \item loosely-coupled VIO and UWB corrections using an EKF.
\end{enumerate}

The MILUV development kit employs VINS-Fusion\footnote{\url{https://github.com/HKUST-Aerial-Robotics/VINS-Fusion}} as a VIO benchmark, which utilizes the calibrated IMU and stereo-camera information to localize. \sh{By default, the VIO estimates are resolved in the frame at which the VINS-Fusion algorithm is initialized. Therefore, the VIO estimates are post-processed to ensure that they are expressed with respect to a common known reference frame, which is chosen as the Vicon frame, $\mc{F}_a$.} The UWB information, height measurements, bottom-facing camera, and the magnetometer are not \jln{used} here, and each robot runs its own VINS-Fusion algorithm independent\jln{ly} of its neighbours. The position estimation accuracy with and without loop closures are reported in Table \ref{tab:vins}. \ijrr{The results are reported for all experiments except for those that did not involve a significant amount of motion. Due to the enclosed space, the localization performance is better when loop closures are enabled; however, there is significant room for improvement for pure visual-inertial odometry, as some of the experiments drift more than 50 cm despite the experiments being relatively short. In particular, experiments \texttt{default\_3\_movingTriangle\_0b} and \texttt{default\_1\_remoteControlLowPace\_0} failed, and would be interesting examples for new algorithms to try and address. Additionally, it appears that experiment \texttt{noAprilTags\_1\_remoteControlHighPace\_2b} is interesting to study, as the high pace and small number of visual features leads to inferior performance, with over a meter of drift.}

\begin{table*}
\scriptsize\sf\centering
\caption{\ijrr{Position root-mean-square errors (RMSEs) in meters obtained from VINS-Fusion using the stereo camera and the IMU.}}
\begin{tabular}{c|ccc|ccc}
& \multicolumn{3}{c}{\textbf{Without Loop Closure}} & \multicolumn{3}{c}{\textbf{With Loop Closure}} \\
\cmidrule(lr){2-4}\cmidrule(lr){5-7}
\textbf{Exp. ID} & \textbf{ifo001} & \textbf{ifo002} & \textbf{ifo003} & \textbf{ifo001} & \textbf{ifo002} & \textbf{ifo003}\\
\midrule
default\_3\_zigzag\_0 & 0.49798 & 0.29481 & 0.24563 & 0.04689 & 0.06173 & 0.06529  \\
default\_3\_zigzag\_1 & 0.48900 & 0.31542 & 0.25276 & 0.05960 & 0.08057 & 0.05862  \\
default\_3\_zigzag\_2 & 0.59707 & 0.19387 & 0.27562 & 0.04963 & 0.09482 & 0.06365  \\
default\_3\_random\_0 & 0.18477 & 0.18739 & 0.11430 & 0.05632 & 0.07050 & 0.04017  \\
default\_3\_random\_0b & 0.13755 & 0.28527 & 0.11525 & 0.04678 & 0.06776 & 0.03279  \\
default\_3\_random2\_0 & 0.38991 & 0.34673 & 0.43581 & 0.04076 & 0.04509 & 0.05834  \\
default\_3\_random3\_0b & 0.32644 & 0.22271 & 0.25752 & 0.04764 & 0.04383 & 0.05415  \\
default\_3\_random3\_1 & 0.37456 & 0.20054 & 0.97147 & 0.04025 & 0.04404 & 0.08210  \\
default\_3\_random3\_2 & 0.32587 & 0.36962 & 0.25421 & 0.04550 & 0.04452 & 0.04871  \\
default\_3\_movingTriangle\_0b & 0.22210 & 0.19092 & Diverged & 0.02123 & 0.02712 & Diverged \\
oneTagPerRobot\_3\_random\_0 & 0.09400 & 0.22016 & 0.28575 & 0.04249 & 0.06275 & 0.07592  \\
noAprilTags\_3\_random3\_2b & 0.28042 & 0.36604 & 0.37193 & 0.05396 & 0.05902 & 0.06957  \\
cir\_3\_random3\_0 & 0.23797 & 0.23220 & 0.28489 & 0.03788 & 0.04157 & 0.04331  \\ 
cir\_3\_random3\_1 & 0.20374 & 0.15422 & 0.20759 & 0.04066 & 0.04545 & 0.06052  \\
cir\_3\_random3\_2 & 0.19251 & 0.34587 & 0.22279 & 0.04640 & 0.04834 & 0.05127  \\
cirObstacles\_3\_random\_0 & 0.11300 & 0.22001 & 0.09569 & 0.03852 & 0.07460 & 0.03007  \\
\midrule
default\_1\_random3\_0 & 0.27893 & - & - & 0.04198 & - & -  \\
default\_1\_random3\_1 & 0.15202 & - & - & 0.03897 & - & -  \\
default\_1\_random3\_2 & 0.16122 & - & - & 0.04180 & - & -  \\
default\_1\_circular2D\_0 & 0.42246 & - & - & 0.06397 & - & -  \\
default\_1\_circular3D\_0 & 0.31271 & - & - & 0.04051 & - & -  \\
default\_1\_remoteControlHighPace\_0 & 0.43561 & - & - & 0.03237 & - & -  \\
default\_1\_remoteControlHighPace\_1 & 0.31542 & - & - & 0.08057 & - & -  \\
default\_1\_remoteControlHighPace\_2 & 0.27008 & - & - & 0.04177 & - & -  \\
default\_1\_remoteControlLowPace\_0 & Diverged & - & - & Diverged & - & - \\
default\_1\_remoteControlLowPace\_0\_v2 & 0.50296 & - & - & 0.04889 & - & -  \\
oneTagPerRobot\_1\_random3\_0 & 0.21106 & - & - & 0.03551 & - & -  \\
noAprilTags\_1\_random3\_2b & 0.24986 & - & - & 0.05874 & - & -  \\
noAprilTags\_1\_remoteControlHighPace\_2b & 1.08099 & - & & 0.08235 & - & -  - \\
obstacles\_1\_random3\_0b & 0.25292 & - & - & 0.03532 & - & -  \\
cir\_1\_circular2D\_0 & 0.34473 & - & - & 0.03643 & - & -  \\
cirObstacles\_1\_random3\_0 & 0.13452 & - & - & 0.03868 & - & - 
\end{tabular}
\label{tab:vins}
\end{table*}

\begin{table}[h]
\scriptsize\sf\centering
\caption{\ijrr{Position RMSEs in meters using interoceptive measurements from the gyroscope and accelerometer in each robot, and tag-to-anchor and tag-to-tag UWB range measurements between robots. The state includes the robots' poses relative to the global frame $\mc{F}_a$ in $SE_2(3)$ and accelerometer and gyroscope biases of the IMUs in each robot.}}
\begin{tabular}{c|ccc}
\textbf{Exp. ID} & \textbf{ifo001} & \textbf{ifo002}  & \textbf{ifo003}\\
\midrule
default\_3\_zigzag\_0 & 0.11372 & 0.12429 & 0.12099 \\
default\_3\_zigzag\_1 & 0.11588 & 0.12515 & 0.12157 \\
default\_3\_zigzag\_2 & 1.53563 & 1.59871 & 4.60612 \\
default\_3\_random\_0 & 0.12159 & 0.12137 & 0.15042 \\
default\_3\_random\_0b & 0.12630 & 0.13991 & 0.13273 \\
default\_3\_random2\_0 & 0.12120 & 0.12281 & 0.12026 \\
default\_3\_random3\_0b & 0.11718 & 0.19750 & 0.74157 \\
default\_3\_random3\_1 & 0.14363 & 0.39839 & 0.14308 \\
default\_3\_random3\_2 & 0.55328 & 0.46906 & 1.03717 \\
default\_3\_movingTriangle\_0b & 0.12605 & 0.15871 & 0.12617 \\
oneTagPerRobot\_3\_random\_0 & 0.20304 & 0.24431 & 0.30646 \\
noAprilTags\_3\_random3\_2b & 0.24714 & 0.70782 & 0.37348 \\
\midrule
default\_1\_random3\_0 & 0.12541 & - & - \\
default\_1\_random3\_1 & 0.14737 & - & - \\
default\_1\_random3\_2 & 0.14877 & - & - \\
default\_1\_circular2D\_0 & 0.19352 & - & - \\
default\_1\_circular3D\_0 & 0.17514 & - & - \\
default\_1\_remoteControlHighPace\_0 & 0.78438 & - & - \\
default\_1\_remoteControlHighPace\_1 & 0.76667 & - & - \\
default\_1\_remoteControlHighPace\_2 & 0.81323 & - & - \\
default\_1\_remoteControlLowPace\_0 & 0.14266 & - & - \\
default\_1\_remoteControlLowPace\_0\_v2 & 0.12665 & - & - \\
oneTagPerRobot\_1\_random3\_0 & 0.19102 & - & - \\
noAprilTags\_1\_random3\_2b & 0.13901 & - & - \\
noAprilTags\_1\_remoteControlHighPace\_2b & 0.15775 & - & - \\
obstacles\_1\_random3\_0b & 0.14545 & - & -
\end{tabular}
\label{tab:ekf_imu}
\end{table}

\begin{table}[t]
\scriptsize\sf\centering
\caption{\ijrr{Position RMSEs in meters using interoceptive measurements from Visual Inertial Odometry on each robot, and tag-to-anchor and inter-tag UWB range measurements. The state consists of the robots' poses relative to the global frame $\mc{F}_a$ in $SE(3)$. Experiments where VINS diverged are omitted.}}
\begin{tabular}{c|ccc}
\textbf{Exp. ID} & \textbf{ifo001} & \textbf{ifo002}  & \textbf{ifo003}\\
\midrule
default\_3\_zigzag\_0 & 0.09994 & 0.10139 & 0.09604\\
default\_3\_zigzag\_1 & 0.10162 & 0.09705 & 0.09254\\
default\_3\_zigzag\_2 & 0.18607 & 0.20341 & 0.86510\\
default\_3\_random\_0 & 0.09889 & 0.09013 & 0.10520\\
default\_3\_random\_0b & 0.10724 & 0.10381 & 0.10943 \\
default\_3\_random2\_0 & 0.10824 & 0.09397 & 0.09500\\
default\_3\_random3\_0b & 0.10134 & 0.15325 & 0.10278\\
default\_3\_random3\_1 & 0.12235 & 0.25739 & 0.10683\\
default\_3\_random3\_2 & 0.11335 & 0.13415 & 0.10677 \\
oneTagPerRobot\_3\_random\_0 & 0.17011 & 0.13821 & 0.12632\\
noAprilTags\_3\_random3\_2b & 0.10702 & 0.22356 & 0.11173\\
\midrule
default\_1\_random3\_0 & 0.11960 & - & -\\
default\_1\_random3\_1 & 0.12937 & - & -\\
default\_1\_random3\_2 & 0.13349 & - & -\\
default\_1\_circular2D\_0 & 0.18469 & - & -\\
default\_1\_circular3D\_0 & 0.17286 & - & -\\
default\_1\_remoteControlHighPace\_0 & 0.96551 & - & -\\
default\_1\_remoteControlHighPace\_1 & 0.87524 & - & -\\
default\_1\_remoteControlHighPace\_2 & 1.03697 & - & -\\
default\_1\_remoteControlLowPace\_0\_v2 & 0.12482 & - & -\\
oneTagPerRobot\_1\_random3\_0 & 0.18393 & - & -\\
noAprilTags\_1\_random3\_2b & 0.12193 & - & -\\
noAprilTags\_1\_remoteControlHighPace\_2b & 0.12489 & - & - \\
obstacles\_1\_random3\_0b & 0.12965 & - & -
\end{tabular}
\label{tab:ekf_vio}
\end{table}

Meanwhile, the EKF utilizing UWB-inertial data does not use the stereo camera, but rather uses the UWB range data between pairs of tags on different quadcopters and between anchors and tags. The IMU is used to propagate the state, while the UWB data and height measurements from the laser rangefinder are used to correct the state. The state vector involves the poses of each quadcopter as an element of $SE_2(3)$ (\cite{Barrau2015}), and the accelerometer and gyroscope biases. \ijrr{The positioning accuracy for all experiments are reported in Table \ref{tab:ekf_imu}, excluding those without a significant amount of motion and those with CIR data, as the ranging frequency is too low. In this case, the EKF achieves 10-20 cm accuracy in most scenarios, which is not sufficiently accurate for controllers and path planning algorithms in safety-critical applications. Additionally, it appears that the EKF struggles in high-pace experiments done with one of the quadcopters, and there is also a failure in the \texttt{default\_3\_zigzag\_2} and \texttt{default\_3\_random3\_2} experiments. These present clear examples for potential improvements.}

Lastly, an EKF that uses the VINS-Fusion output in a loosely-coupled manner for state propagation rather than the IMU is presented. Translational velocity estimates generated from the VINS-Fusion output and the gyroscope readings for each quadcopter are used to propagate the state, and all the inter-tag and anchor-to-tag UWB range data are used to correct the state. The state vector consists of the robot poses as an element of $SE(3)$. As with the UWB-inertial localization EKF, this approach also uses the EKF height measurements from the laser rangefinder. \ijrr{The results are reported in Table~\ref{tab:ekf_vio}, where the accuracy also falls in the 10-20 cm range and the EKF struggles in high-pace experiments.}



\section{Potential Applications} \label{sec:potential_applications}
The target application of the MILUV dataset is multi-UAV navigation with the option to fuse both UWB and visual measurements in the estimation pipeline. However, due to the variety of sensor information available, MILUV is a multipurpose dataset that can be used in applications beyond multi-robot localization.

\subsection{Relative Pose Estimation} \label{subsec:potential_applications_rpe}
The problem of relative pose estimation between robots is an active area of research due to its importance in multi-robot teams. UWB and vision are particularly useful when each robot is trying to estimate where other robots are relative to themselves. When equipped with cameras, a robot can detect its neighbours and obtain direct relative pose measurements, as is done in \cite{Xu2020a}. Meanwhile, UWB 
range measurements between robots 
also provide information regarding the relative poses. This has been utilized in recent papers (e.g., \cite{Nguyen2018, Charles2021RP, Shalaby2024passive}) to estimate the relative poses between robots, and can be combined with vision measurements to aid in scenarios where the neighbouring robots fall outside of the camera's field-of-view.

\subsection{Visual-Inertial-Range SLAM}

SLAM stands for simultaneous localization and mapping, and is the problem where a robot attempts to construct a map of the environment and localize itself within it. SLAM is typically done using cameras or \sh{LiDAR} as they are better suited for sensing the environment, but due to the difficulties associated with SLAM stemming from the data-association problem, lighting, and the inherent unobservability, UWB has been gaining interest in the SLAM literature (e.g., \cite{Cao2020, Jung2022, Hu2024}). The \sh{MILUV} dataset provides all the data needed to evaluate Visual-Inertial-Range SLAM algorithms, where the UWB anchors can also be assumed to be at unknown locations to be estimated. The AprilTags provide easy-to-detect visual features, and the UWB information can be used to aid with the data-association problem or loop closures. 

Moreover, multi-robot SLAM is yet another active field of research (e.g., \cite{XieSLAM2022, Tian2023}) where MILUV can be utilized. Alongside a relative position estimation algorithm, the different robots can synthetically share their mapping information for inter-robot loop closures and collaborative map building. 


\subsection{CIR-based Localization}

\begin{table}
\small\sf\centering
\caption{The accuracy of the different models trained for LOS-NLOS classification for Experiment \texttt{cirObstacles\_3\_random\_0}.}
\begin{tabular}{l|c}
\textbf{Model} & \textbf{Accuracy} \\
\midrule \\
ExtraTreeClassifier              &  0.69        \\
QuadraticDiscriminantAnalysis    &  0.51        \\
NearestCentroid                  &  0.58        \\
PassiveAggressiveClassifier      &  0.65        \\
NuSVC                            &  0.80        \\
Perceptron                       &  0.63        \\
SGDClassifier                    &  0.72        \\
BaggingClassifier                &  0.78        \\
XGBClassifier                    &  0.78        \\
LinearSVC                        &  0.64        \\
RandomForestClassifier           &  0.80        \\
LGBMClassifier                   &  0.79        \\
CalibratedClassifierCV           &  0.79        \\
LabelSpreading                   &  0.79        \\
SVC                              &  0.79        \\
ExtraTreesClassifier             &  0.79        \\
DummyClassifier                  &  0.79        \\
LabelPropagation                 &  0.79        \\
LogisticRegression               &  0.65        \\
AdaBoostClassifier               &  0.68        \\
BernoulliNB                      &  0.65        \\
GaussianNB                       &  0.74        \\
RidgeClassifier                  &  0.59        \\
KNeighborsClassifier             &  0.70        \\
RidgeClassifierCV                &  0.60        \\
DecisionTreeClassifier           &  0.62        \\
LinearDiscriminantAnalysis       &  0.56
\end{tabular}
\label{tab:los_classification}
\end{table}

The CIR data incorporates a lot of information regarding the environment of a robot, 
and therefore can unlock a whole different approach for localization and mapping. 
The main challenge of inferring \jln{environmental} information from the CIR is that 
it is hard to analytically model the behaviour of the response, as detections of reflections 
\jln{from} different objects in the environment \jln{are superimposed},
different materials reflect differently, and the 
\jln{measured} response varies in LOS and NLOS conditions. 

Nonetheless, it is possible that with sufficient data, models can be trained for 
environment inference using 
CIR data. A basic example using the \texttt{lazypredict}\footnote{\url{https://pypi.org/project/lazypredict/}} 
library in Python for LOS-NLOS classification is provided in the MILUV devkit, 
and the results for different classifiers are shown in Table \ref{tab:los_classification}.

There has also been interest in the literature on CIR-based localization and mapping \citep{Leitinger2019, Ledergerber2020, Ninnemann2021}. However, current approaches are typically limited in scope as data collection using mobile robots and CIR data is a challenging task and there are no extensive publicly-available datasets that include CIR data for multiple robots. By providing CIR data alongside data from many other sensors for multiple robots, MILUV allows the development of CIR-based solutions for resource-constrained systems in challenging environments.


\section{Limitations and Possible Extensions} \label{sec:limitations_and_possible_extensions}
Before concluding this paper, it is worth mentioning some known limitations and potential areas of improvement for this dataset. The subsequent list is only accurate at the time of writing. The GitHub repository associated with MILUV will continue to be updated and maintained by the authors of this paper and hopefully the community, meaning some of the subsequent points might be addressed at the time of reading. 

\begin{itemize}
    \item \textbf{Lack of outdoor experiments}: The experiments that comprise the MILUV dataset are performed indoors, which imposed environmental limitations on data collection. Due to the relatively small size of the flight arena, a maximum of three UAVs could be flown per experiment at a maximum speed of $\SI{4.418}{\meter/\second}$. The small flight arena also prohibited the safe execution of experiments with dynamic obstacles. A complimentary outdoor version of the MILUV dataset would allow for longer and faster flight trajectories with the ability to add more UAVs, other robotic platforms, and dynamic obstacles to the multi-robot scenarios. 
    \item \textbf{Biased height measurements}: It appears that the height measurements available from the laser rangefinder are biased as compared to the ground truth, probably due to the motion capture system's frame not being exactly at the ground. As such, a bias correction must be applied to the height measurements before using them by comparing them to the ground truth data. \ijrr{The biases for the 3 quadcopters are provided in the development kit in \texttt{config/height/bias.yaml}, and the data loader in the development kit allows for automatically correcting this bias in the rangefinder data.}
    \item \textbf{Low frequency of the CIR data}: The CIR involves much more data than just range measurements, and are thus only collected at a low rate. Whereas the ranging data are collected at a rate of up to $200 \,\si{Hz}$ for most experiments, experiments with CIR data have UWB data only collected at about $5\,\si{Hz}$.
    \item \textbf{Missing information from the ground truth}: This dataset can be used in the estimation of the clock states (offset and skew) of the UWB transceivers relative to one another as part of the localization algorithms (e.g., \cite{Shalaby2024passive}). Nonetheless, clock states are not ``measurable'', and are thus not provided as part of the ground truth, complicating the evaluation and design process of such algorithms. The evaluation would have to rely on the accuracy of the pose estimates in the dataset and the IMU biases provided by the development kit as discussed in Section \ref{subsec:imu_calib}, but an interesting approach would be to provide the clock states and maybe the IMU biases achieved through a batch optimization solution.
\end{itemize}

\section{Conclusion}

This paper \sh{presents} MILUV, a multi-UAV indoor localization dataset using UWB and vision measurements. MILUV is unique in that it includes multiple robots, low-level UWB data such as the raw timestamps and CIR, and vision data from a calibrated stereo camera. Three Uvify IFO-S quadcopters, each equipped with two custom-made UWB transceivers, a stereo camera, two IMUs, a magnetometer, a downward-facing camera, a barometer, and a downward-facing laser rangefinder are flown indoors. The environment includes 6 static UWB anchors, 48 April Tags, 
and a Vicon motion capture system to record the
\jln{true} pose of the robots. The presented dataset includes many different experiments with varying combinations of number of robots, number of UWB tags, environments, and motion profiles. Calibration and obstacle data are also provided. 

Alongside the provided data, a development kit is provided to facilitate the use of MILUV for the evaluation of algorithms. Various examples are provided, such as VIO and standard EKF algorithms, that provide an introduction into how the development kit can be used as well as benchmarks to be used as comparison points for novel algorithms developed by the users of this dataset. The provided development kit is made open-source, and will continue to be maintained by the authors of this paper, and hopefully by the community. 

\begin{funding}
This work was supported by the NSERC Discovery and Alliance Grant programs, the Canadian Foundation for Innovation (CFI) program, and the Denso Corporation.
\end{funding}

\begin{acks}
The authors would like to thank Saad Chidami for his assistance in the development of the custom UWB modules, as well as Steven Dahdah for his assistance with the Uvify hardware.
\end{acks}


\bibliographystyle{SageH}
\bibliography{ref.bib}

\end{document}